\documentclass[10pt,twocolumn]{article}
\usepackage[letterpaper,margin=0.75in]{geometry} %
     
\usepackage{graphicx}
\usepackage{authblk}   
\graphicspath{{figures/}} %
\usepackage{amsmath} %
\usepackage{amssymb} %
\usepackage{booktabs} %
\usepackage[round]{natbib} \AtBeginDocument{\setlength{\bibsep}{0.6\bibsep}} %
\usepackage[hyphens]{url} %
\makeatletter
\renewcommand\section{\@startsection{section}{1}{\z@}{-3.5ex \@plus -1ex \@minus -.2ex}{2.3ex \@plus.2ex}{\normalfont\Large\bfseries\raggedright}}
\renewcommand\subsection{\@startsection{subsection}{2}{\z@}{-3.25ex\@plus -1ex \@minus -.2ex}{1.5ex \@plus .2ex}{\normalfont\large\bfseries\raggedright}}
\makeatother
\usepackage{float} %
\newif\ifanonymous \anonymousfalse
\usepackage[hidelinks]{hyperref} %

\title{FireGraphNet: Sub-daily wildfire spread prediction \\ with directed graph neural networks} %
\ifanonymous\author{}\else
\author[1,*]{Aiyshwariya Paulvannan Kanmani}
\author[1]{Miguel Esparza}
\author[1]{Ali Mostafavi}
\affil[1]{Urban Resilience.AI Lab, Zachry Department of Civil and Environmental Engineering, Texas A\&M University, College Station, TX 77840, United States of America}
\affil[*]{Corresponding author: aiyshwariya@tamu.edu}
\fi
\date{}
\newcommand{\FGN}{FireGraphNet}
\newcommand{\RelatedRef}{Supplementary Note~\ref{sec:related}} %
\makeatletter
\newcommand{\numlabel}[2]{{\protected@edef\@currentlabel{\csname the#1\endcsname}\label{#2}}} %
\newcommand{\numlabelafter}[2]{{\protected@edef\@currentlabel{\csname the#1\endcsname}\edef\@currentHref{#1.\csname theH#1\endcsname}\label{#2}}} %
\newcommand{\EDnumbering}{\setcounter{figure}{0}\setcounter{table}{0}\renewcommand{\thefigure}{\arabic{figure}}\renewcommand{\thetable}{\arabic{table}}%
  \renewcommand{\theHfigure}{ED.\arabic{figure}}\renewcommand{\theHtable}{ED.\arabic{table}}%
  \renewcommand{\figurename}{Extended Data Fig.}\renewcommand{\tablename}{Extended Data Table}}
\newcommand{\SuppTablenumbering}{\setcounter{table}{0}\renewcommand{\thetable}{\arabic{table}}\def\p@table{}%
  \renewcommand{\theHtable}{S.\arabic{table}}\renewcommand{\tablename}{Supplementary Table}}
\makeatother
\newcounter{suppnote}
\newcommand{\suppnotesection}[1]{\refstepcounter{suppnote}\subsection*{Supplementary Note \thesuppnote: #1}} %
\begin{document}
\maketitle
\begin{abstract}
Wildfire decisions are made in hours, yet most learned spread models predict daily on targets including burned area. We propose \FGN{} which formulates 12-hour growth as node classification on directed H3 graphs, with candidates ahead of the front and wind alignment, terrain, fuel continuity and barriers on edges. On satellite perimeters it achieves average precision 1.7--1.9$\times$ the strongest distance baseline on test California fires and 1.62$\times$ out of state. We show this skill comes from message passing rather than cell features or the aggregator, is highest on young fires in state, is carried deep along the wind by its edges, and is hidden by pooling. A convolutional network matches this ranking once trained on new growth, at worse calibration, so candidate construction, target definition, and aggregation shape conclusions about model skill. Editing fuel and barrier attributes reduces predicted spread more under guided than random placement, extending \FGN{} to operational intervention analysis. %

\end{abstract}
\section{Introduction} \label{sec:intro} %
Wildfire response requires anticipating where an active fire may advance during the next several hours as changes in its position can alter evacuation planning, crew access, and the placement of protective resources. An observed perimeter identifies the area already affected, but decisions require an assessment of the adjacent landscape that may burn next. Daily summaries can obscure the timing and sequence of growth within that decision window, even when they describe the same total burned area (Extended Data Figure~\ref{fig:edsubdaily}). Satellite-derived perimeter sequences such as the Fire Event Data Suite (FEDS) provide an opportunity to study growth at approximately 12-hour intervals using observations of actual fires \citep{chen2022feds}. Translating these observations into useful predictions requires a formulation that connects the evolving fire front to the terrain, fuels, and atmospheric conditions surrounding it. The central challenge is to determine where those spatial interactions yield predictive information beyond the expectation that fire usually advances into nearby cells.

Spread is inherently relational: the relevance of wind, terrain, and fuel depends on the direction and connection between an already burning location and the landscape ahead of it. A directed graph offers a natural representation of these relationships by attaching directional attributes to the links between neighboring cells. Its value, however, must be established for a specific forecasting target and candidate population, since a model can receive credit for reproducing existing burned area or benefit from a candidate region defined using future observations. These considerations motivate a study that couples an observation-derived, sub-daily prediction task with controlled comparisons and explicit spatial baselines. Such a design can reveal both the predictive value of the graph formulation and the circumstances in which its apparent advantages depend on how performance is measured.

\begin{figure*}[!tp] \centering \includegraphics[width=\textwidth]{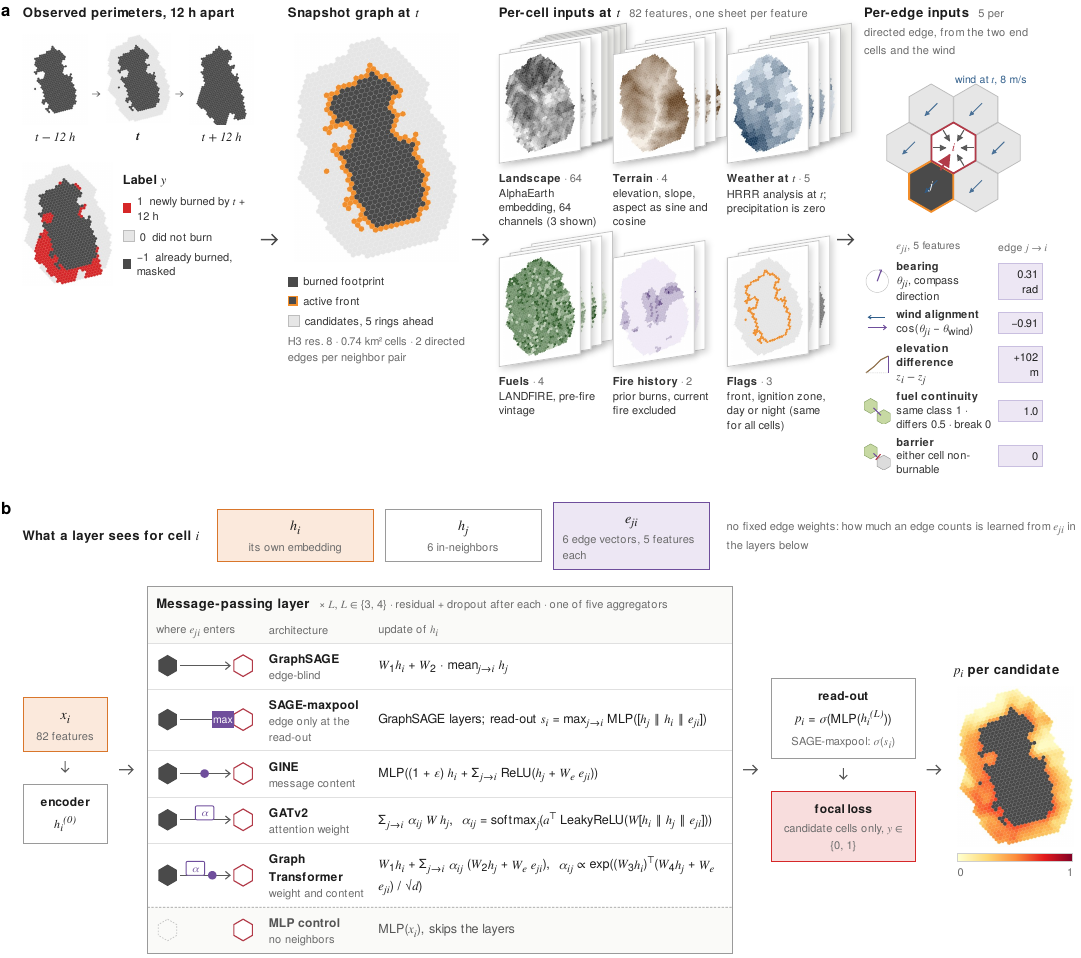} \caption{The \FGN{} formulation illustrated on the Creek Fire snapshot of 6 September 2020, 12:00 UTC. (a) From perimeters to one snapshot. The perimeter at $t - 12$\,h shows the cadence and is not a model input. The perimeter at $t + 12$\,h is used only for the label, which is 1 for a candidate cell that newly burned, 0 for one that did not, and $-1$ for already burned cells, which are masked. At $t$, the graph shows the burned footprint, the active front and the candidate cells within 5 rings of the front. The 82 per-cell inputs are drawn one sheet per feature, grouped by source. The 5 per-edge inputs are shown for one candidate cell $i$ with its 6 directed in-edges, the HRRR wind at $t$ and the edge vector from the burning neighbor $j$. (b) The architecture. A message-passing layer receives, for cell $i$, its own embedding, the embeddings of its 6 in-neighbors and their 6 edge vectors. The five rows are the five aggregators, ordered by how much of the edge vector enters the computation (Section~\ref{sec:models}), and the MLP control skips the layers. The read-out returns one probability per candidate cell, shown for GINE.} \label{fig:formulation} \end{figure*} %

Existing approaches to wildfire spread prediction fall broadly into physics-based simulation and learned prediction (\RelatedRef). Most learned models are convolutional neural networks (CNNs) trained on raster maps such as the Next Day Wildfire Spread (NDWS) and WildfireSpreadTS (WSTS) datasets \citep{andrianarivony2024, huot2022, gerard2023}. The closest prior work, a node-classification study of observed perimeter spread on graphs \citep{roesch2024}, works at daily cadence with undirected, feature-less edges and scores cells drawn from the fire\textquoteright s final perimeter. Graph formulations \citep{jiang2022, roesch2024} are of particular interest here because raster maps or image-based models are restrictive in what they can represent at the fire front. A graph instead carries directional quantities on its edges, and lets barriers and fuel breaks be expressed as edits to those edges.

Widely used wildfire learning datasets frame spread prediction as a next-day raster task, while graph-based studies have demonstrated the feasibility of representing observed fire progression through connected landscape cells \citep{huot2022, gerard2023, roesch2024}. Recent FIRE-AID work also describes sub-daily prediction from satellite-derived perimeters with graph-based message passing, although no quantitative evaluation has yet been published \citep{laux2026fireaid, esa2026fireaid}. The research question therefore extends beyond whether graphs can predict spread to how directional landscape interactions contribute to short-horizon skill under a clearly defined prediction task. Several studies have observed that next-day raster models lean on the previous fire mask, and that scoring it as part of the target inflates the reported skill \citep{huot2022, firesensenet, esparza2026modular}. To our knowledge, no study has tested whether the training objective is itself the cause, i.e., whether a model trained and scored without the persisting footprint learns new growth beyond a distance rule. A useful assessment must distinguish newly burned cells from the existing footprint, construct the candidate region without knowledge of future growth, and determine what the learned model adds beyond simple spatial rules. It must also establish whether conclusions persist across individual fires and distances from the front, because aggregate performance can conceal failures in the conditions that matter for response. \FGN{} addresses these questions through approximately 12-hour node classification on directed H3 graphs, combining physically motivated edge attributes with explicit spatial baselines, objective-matched comparisons, and geographic transfer tests. Its point of departure is the integrated formulation and empirical characterization of near-front spread skill under these constraints. %

\begin{figure*}[!tp] \centering \includegraphics[width=\textwidth]{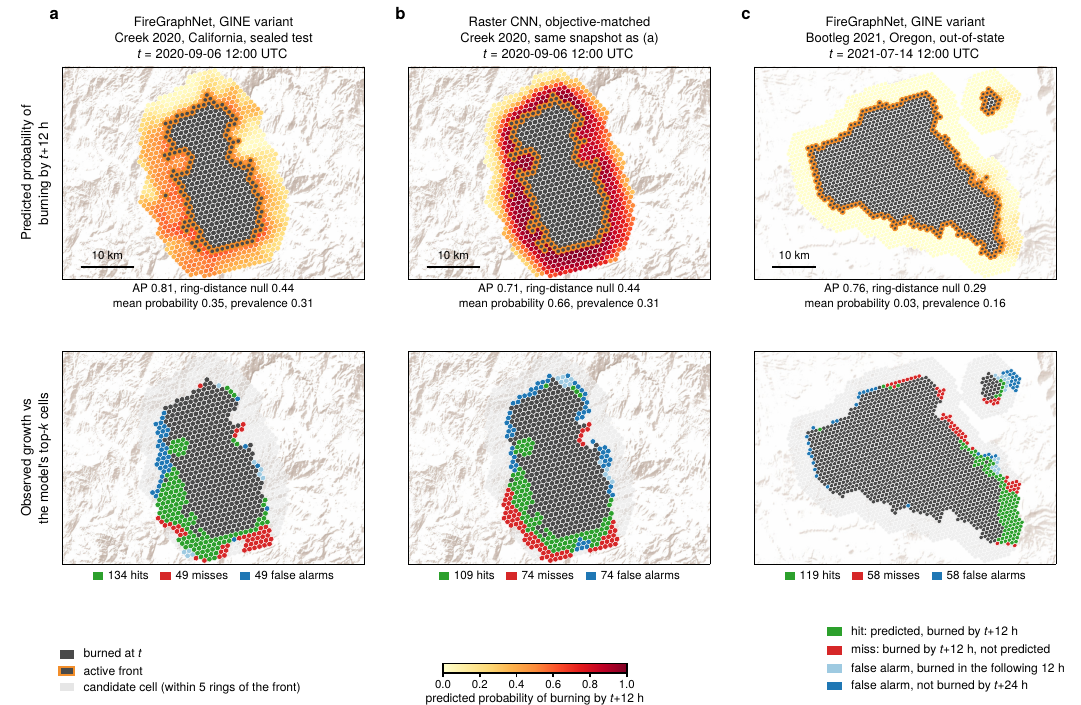} \caption{Three 12-hour predictions, each the mean over 10 seeds. (a) The GINE variant on Creek 2020 (California, sealed test), 6 September 2020, 12:00 UTC%
. (b) The objective-matched raster CNN on the same snapshot. (c) The GINE variant on Bootleg 2021 (Oregon, sealed out-of-state set), 14 July 2021, 12:00 UTC. Top row: the burned footprint (dark), the active front (orange) and each candidate cell within 5 rings of the front colored by its uncalibrated probability of newly burning by $t + 12$\,h. Bottom row: observed growth against the model's $k$ highest-ranked cells, where $k$ is the number of cells that burned. False alarms are split by whether the cell burned in the following 12 hours. The CNN field is saturated (mean 0.66, prevalence 0.31) where GINE's is graded (0.35), which is the miscalibration reported in Section~\ref{sec:results-raster}. 
GINE ranks higher on this snapshot (AP 0.81 vs 0.71) but the CNN is higher over the whole Creek fire (0.1210 vs 0.1073, Supplementary Table~\ref{tab:edperfire}), so (b) illustrates calibration rather than a ranking advantage. On Bootleg the probabilities are low (0.03 vs prevalence 0.16) yet the ranking still finds the eastern flank. Basemap tiles: Esri World Terrain.} \label{fig:predmap} \end{figure*}

\FGN{} poses 12-hour spread prediction as a question about the cells around the fire. The landscape is divided into hexagonal cells of about 1\,km from the global H3 grid, so every cell has 6 neighbors and the grid is the same for every fire. Each join between neighboring cells is a directed edge that carries how well the wind aligns with it, how much the ground climbs along it and whether the fuel continues across it. A graph neural network (GNN) then updates the description of every cell from its neighbors along these edges, which is called message passing. The rule that combines a cell's incoming messages is called its aggregator. We train five message-passing architectures within this formulation (GraphSAGE, SAGE-maxpool, GINE, GATv2 and a Graph Transformer), which we call its variants. Wherever a single model is needed we use the GINE variant (a graph isomorphism network with edge features), since it is the best calibrated of the five. For each candidate cell, i.e., an unburned cell within 5 rings of the front (the burned cells that touch an unburned cell), \FGN{} returns the probability that the cell newly burns by $t + 12$\,h, where $t$ is the time of the latest observed perimeter (Figure~\ref{fig:formulation}). In practice, \FGN{} takes the latest perimeter from a satellite tracking service such as FEDS, with the weather analysis valid at that snapshot's nominal time stamp, and returns a ranked map of the cells most at risk over the next 12 hours (Figure~\ref{fig:predmap}). Every input is public, and no forecast product is needed.  %

In this study, we examine whether a directed graph representation of the active fire front and its surrounding landscape can support 12-hour prediction of newly burned cells, and identify the conditions under which its measured skill is informative. RQ1 quantifies the contribution of learned neighborhood context beyond deterministic spatial rules and the tested cellwise control, and assesses differences among message-passing architectures. RQ2 characterizes how predictive skill varies with fire age, distance from the front, and weather information, so that a pooled result can be read against distinct spread conditions. RQ3 compares graph and raster formulations under a common candidate objective and evaluates transfer of the graph model to fires outside its training region. RQ4 tests whether driver-type conditioning and auxiliary intensity or longer-horizon supervision improve performance across fires rather than only within a snapshot-weighted aggregate. An additional exploratory analysis evaluates responses to edited fuel and barrier attributes, connecting the representation to intervention hypotheses without assuming that model responses establish treatment effectiveness. Together, these objectives connect the proposed representation to interpretable evidence about its skill, scope, and potential use. %

This study contributes a directed graph formulation and a systematic characterization of approximately 12-hour wildfire growth prediction from satellite-derived perimeter sequences. Burned cells provide spatial context, unburned cells within five rings of the current front define the prediction population, and edge attributes represent wind alignment, elevation differences, fuel continuity, and barriers. Across five California test fires, the five \FGN{} variants achieve average precision approximately 1.7--1.9$\times$ that of the strongest spatial reference rule; the GINE variant retains a 1.62-fold advantage over the distance baseline on five fires outside California. Skill is highest on young fires and falls with track age for every variant and for the distance rule, so pooled scores are dominated by old-fire snapshots. A CNN given the same node features plus wind components shows why the training objective matters. Trained on the burned footprint it ranks new growth below the distance baseline, whereas supervised on the same candidate cells it reaches a ranking statistically indistinguishable from the GINE variant, with worse calibration. The analysis further shows that weather contributions depend on distance from the front: masking weather improves the pooled snapshot statistic while degrading deeper-ring predictions, demonstrating that aggregation can obscure environmentally relevant skill. Comparisons across variants and training objectives identify differences in raw calibration and threshold behavior, while the available evidence does not resolve a ranking advantage between the GINE variant and the objective-matched CNN. Together, these findings characterize the contribution of learned spatial context and demonstrate how candidate construction, target definition, and aggregation determine the interpretation of spread-model skill. %

\section{Results} \label{sec:results}

\begin{figure*}[!tp] \centering \includegraphics[width=\textwidth]{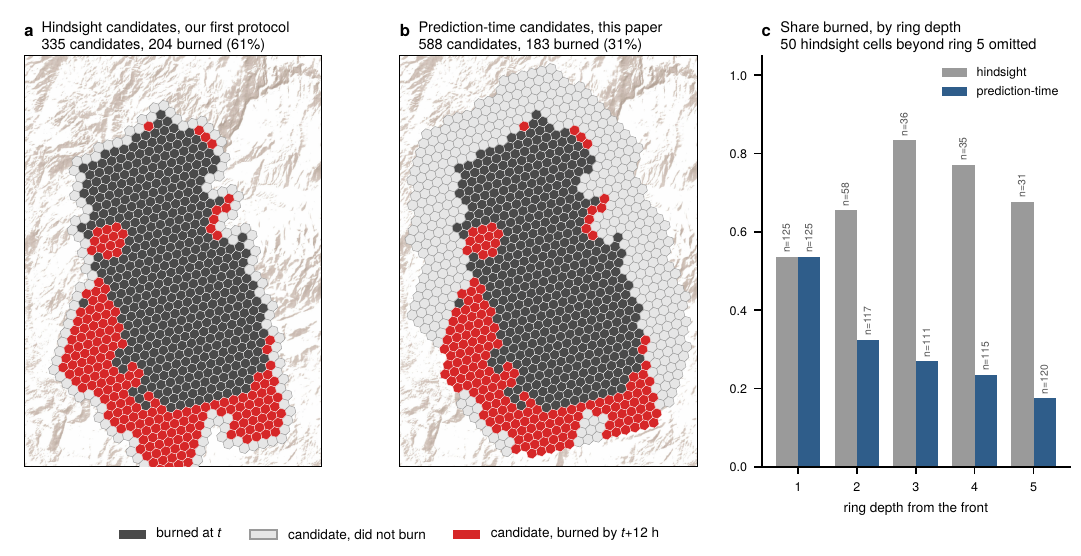} \caption{The Creek 2020 snapshot of Figure~\ref{fig:formulation} under two candidate constructions. (a) Our first protocol, which built candidates as the closest prior work does \citep{roesch2024}, flooding outward through the cells that actually burned: 335 cells, 204 burned (61\%), 21 of them beyond 5 rings of the front. (b) The prediction-time construction of this paper: the 588 unburned cells within 5 rings of the front, 183 burned (31\%). (c) The share of candidates that burned at each ring depth, with candidate counts above the bars. The 50 hindsight candidates beyond ring 5 (21 burned) are not shown. At depth 1 the two sets are identical. Beyond it, the hindsight positive rate is higher at every deeper ring (0.66--0.83 on this snapshot), because deep candidates existed only where the fire went, while the prediction-time rate falls (Supplementary Note~\ref{sec:v1}). Basemap tiles: Esri World Terrain.} \label{fig:candsets} \end{figure*}

Every score is average precision (AP, also written PR-AUC as the area under the precision--recall curve), which measures how well the cells that burned are ranked above those that did not, and a model with no skill scores the prevalence, i.e., the share of candidate cells that burned (Figure~\ref{fig:candsets}b). Lift is the ratio of a model's AP to that of the strongest null model, where a null model is a simple reference rule that needs no learning. Pooled scores use all candidates at once and per-fire scores average over fires, while a stratum is one ring depth or one band of track age (time since first detection), and a snapshot is one fire at one perimeter time $t$. Calibration is whether a predicted probability matches the observed burn frequency, and is summarized by the Brier score and the expected calibration error (ECE), where lower is better (Section~\ref{sec:metrics}). The critical success index (CSI) is the hit rate at a fixed warning threshold $\tau$, reported at the threshold frozen on the validation fires (CSI@$\tau$) and at the best threshold (CSI$_{\max}$).
\subsection{Neighborhood context carries the 12-hour skill} \label{sec:results-headline} %

\begin{table*}[t]
\centering
\caption{Pooled sealed in-state test results: 12\,h spread prediction on the identical
candidate population (512,742 candidates, 8,455 positives, prevalence
0.0165). Mean $\pm$ std over 10 seeds; nulls are deterministic (single pass).
CSI@$\tau$ uses the validation-frozen threshold and CSI$_{\max}$ the best threshold. Objective-matched = trained on the same candidate cells as the graph variants; full-mask = trained on every cell burned by $t+12$\,h, including the existing footprint (Section~\ref{sec:results-raster}). All probabilities are raw
(uncalibrated).}
\label{tab:main}
\small
\setlength{\tabcolsep}{4pt}
\resizebox{\textwidth}{!}{%
\begin{tabular}{lccccc}
\toprule
Model & PR-AUC & CSI@$\tau$ & CSI$_{\max}$ & Brier & ECE \\
\midrule
SAGE, edge-blind (graph) & 0.0624 $\pm$ 0.0064 & 0.0579 $\pm$ 0.0067 & 0.0609 $\pm$ 0.0065 & 0.0945 $\pm$ 0.0177 & 0.2204 $\pm$ 0.0335 \\
SAGE-maxpool (graph) & 0.0675 $\pm$ 0.0085 & 0.0660 $\pm$ 0.0080 & 0.0679 $\pm$ 0.0064 & 0.0638 $\pm$ 0.0078 & 0.1910 $\pm$ 0.0197 \\
GINE (graph) & 0.0710 $\pm$ 0.0056 & 0.0670 $\pm$ 0.0061 & 0.0685 $\pm$ 0.0053 & 0.0348 $\pm$ 0.0076 & 0.0925 $\pm$ 0.0164 \\
GATv2 (graph) & 0.0668 $\pm$ 0.0050 & 0.0624 $\pm$ 0.0064 & 0.0646 $\pm$ 0.0056 & 0.0597 $\pm$ 0.0098 & 0.1472 $\pm$ 0.0218 \\
Graph Transformer (graph) & 0.0638 $\pm$ 0.0078 & 0.0620 $\pm$ 0.0093 & 0.0653 $\pm$ 0.0079 & 0.0500 $\pm$ 0.0075 & 0.1634 $\pm$ 0.0238 \\
MLP (no message passing) & 0.0394 $\pm$ 0.0015 & 0.0354 $\pm$ 0.0025 & 0.0363 $\pm$ 0.0017 & 0.0928 $\pm$ 0.0065 & 0.2710 $\pm$ 0.0127 \\
Res18-U-Net, \emph{objective-matched} (raster) & 0.0769 $\pm$ 0.0134 & 0.0534 $\pm$ 0.0113 & 0.0771 $\pm$ 0.0108 & 0.2394 $\pm$ 0.0663 & 0.3290 $\pm$ 0.0828 \\
Res18-U-Net, \emph{full-mask} (raster, burned-footprint target) & 0.0230 $\pm$ 0.0013 & 0.0169 $\pm$ 0.0042 & 0.0258 $\pm$ 0.0016 & 0.1379 $\pm$ 0.0204 & 0.2426 $\pm$ 0.0266 \\
\midrule
Null: ring-distance (near-first) & 0.0369 & --- & 0.0434 & --- & --- \\
Null: wind--slope heuristic & 0.0182 & --- & 0.0182 & --- & --- \\
Null: ring-distance (deep-first, anti-null) & 0.0125 & --- & 0.0165 & --- & --- \\
\midrule
No skill (prevalence) & 0.0165 & --- & --- & --- & --- \\
\bottomrule
\end{tabular}}
\end{table*}

Table~\ref{tab:main} reports the results on the 5 sealed test fires (used for no model selection or tuning) over 10 seeds, with the per-fire decomposition in Supplementary Table~\ref{tab:edperfire} and Supplementary Note~\ref{app:mechanism}. Two key observations are noted for RQ1. First, \textbf{cell features alone carry no pooled skill beyond the distance rule}. The untuned control, a multilayer perceptron (MLP) on cell features without message passing, scores AP 0.0394, which is the level of the ring-distance null (cells ranked by ring distance, nearest first, at 0.0369). Every variant, in contrast, clears the strongest null by 1.7--1.9$\times$, which is 3.8--4.3$\times$ the prevalence. Second, \textbf{no variant beats another on ranking}. Every pairwise contrast against the best-ranked variant, which is GINE, is indistinguishable after Holm correction (all $p \geq 0.078$; Extended Data Figure~\ref{fig:edsealed}c--e). Since the differences between the variants (0.003--0.009 AP) are below what 5 test fires can detect (Section~\ref{sec:metrics}), we do not read the tie as equivalence.  The one model that never sees the edge features (edge-blind GraphSAGE) has the lowest pooled AP of the 5 and the worst Brier and ECE. The edge-conditioned GINE variant is the best calibrated by a wide margin, with a Brier score of 0.035 against 0.050--0.094 for the other variants and an ECE of 0.093 against 0.147--0.220 (Table~\ref{tab:main} and Extended Data Figure~\ref{fig:edcalibration}). This ECE ordering also reproduces on the 11-fire development pool (Supplementary Table~\ref{tab:calibration}). %
\subsection{The skill transfers to out-of-state fires} \label{sec:results-w2} %

\begin{table*}[t]
\centering
\caption{Pre-registered out-of-state transfer: pooled and stratified PR-AUC
of the GINE variant at its selected configuration, frozen before the run (locked), on the sealed 5-fire out-of-state set (194,796 candidates,
5,952 positives, prevalence 0.0306), evaluated exactly once. Lift is
model over the ring-distance null. The MLP control (no message passing; the
same untuned configuration as in Table~\ref{tab:main}) is shown pooled
only.}
\label{tab:w2}
\small
\setlength{\tabcolsep}{4pt}
\begin{tabular}{lccc}
\toprule
Stratum & PR-AUC & Null & Lift \\
\midrule
Pooled & 0.0964 $\pm$ 0.0057 & 0.0595 & 1.62$\times$ \\
Pooled, MLP (no message passing) & 0.0632 $\pm$ 0.0019 & 0.0595 & 1.06$\times$ \\
\midrule
\quad Santiam Complex 2020 (OR, wind) & 0.1336 $\pm$ 0.0100 & 0.0415 & 3.22$\times$ \\
\quad Holiday Farm 2020 (OR, wind) & 0.1447 $\pm$ 0.0150 & 0.0839 & 1.72$\times$ \\
\quad Cold Springs--Pearl Hill 2020 (WA, grass/wind) & 0.0470 $\pm$ 0.0128 & 0.0337 & 1.39$\times$ \\
\quad East Troublesome 2020 (CO, wind+beetle-kill) & 0.1099 $\pm$ 0.0115 & 0.0978 & 1.12$\times$ \\
\quad Bootleg 2021 (OR, fuel/drought) & 0.1296 $\pm$ 0.0112 & 0.0941 & 1.38$\times$ \\
\midrule
\quad Established track (age $\geq$ 12\,h) & 0.0992 $\pm$ 0.0062 & 0.0575 & 1.73$\times$ \\
\quad Initial window (age $<$ 12\,h) & 0.4590 $\pm$ 0.0172 & 0.5366 & 0.86$\times$ \\
\quad Wind-dominated snapshots & 0.1551 $\pm$ 0.0070 & 0.1023 & 1.52$\times$ \\
\quad Fuel-dominated snapshots & 0.0627 $\pm$ 0.0099 & 0.0465 & 1.35$\times$ \\
\midrule
\quad Track age 0--48\,h & 0.1737 $\pm$ 0.0098 & 0.1714 & 1.01$\times$ \\
\quad Track age 48--240\,h & 0.1154 $\pm$ 0.0188 & 0.0630 & 1.83$\times$ \\
\quad Track age 240--480\,h & 0.0661 $\pm$ 0.0129 & 0.0589 & 1.12$\times$ \\
\quad Track age 480\,h+ & 0.1387 $\pm$ 0.0120 & 0.0367 & 3.78$\times$ \\
\bottomrule
\end{tabular}
\end{table*}

The single pre-registered out-of-state pass answers the transfer half of RQ3, i.e., whether \FGN{}'s skill is a property of the formulation or depends on having seen California landscapes in training. This experiment confirms the transfer claim under the commitments written before the run (Table~\ref{tab:w2}), with a pooled lift of 1.62$\times$ over the ring-distance null (1.92$\times$ in state) that holds in the established-track stratum (1.73$\times$). The no-message-passing control transfers no skill (1.06$\times$, i.e., at the null). The pre-registration named the wind-dominated stratum as the hard case, since a pre-registered check on the 11 non-test fires (the development pool, Section~\ref{sec:corpus}) had found a wind-class skill deficit (Supplementary Note~\ref{supp:enrich}). On this wind-heavy set that stratum holds its lift (1.52$\times$ against 1.35$\times$ fuel-dominated). Per-fire lifts all exceed 1, from Santiam at 3.22$\times$ to East Troublesome at 1.12$\times$ (remaining strata in Supplementary Note~\ref{app:w2}). %
\subsection{\FGN{} outperforms the raster CNN in operation} \label{sec:results-raster} %

This subsection covers the head-to-head half of RQ3 by putting a raster model through the same evaluation as \FGN{}.  Trained on the burned footprint, i.e., full-mask segmentation of every cell burned by $t+12$\,h, the CNN ranks 12\,h spread below the distance null (0.0230 vs 0.0369). This target plays the same role for perimeter labels as the next-day targets of NDWS and WSTS \citep{huot2022, gerard2023}, since both contain the persisting footprint. %

When re-supervised on the candidate objective the graph models use, the same architecture recovers to a level statistically indistinguishable from the best-ranked graph model on pooled ranking (0.0769 $\pm$ 0.0134 vs GINE 0.0710 $\pm$ 0.0056; Welch $p = 0.22$). This second pass was declared before it ran, after the first had been scored (Supplementary Note~\ref{app:contacts}). However, the two formulations separate on \textbf{operating behavior}, i.e, in terms of calibration, threshold transfer and seed stability. The objective-matched CNN is miscalibrated, with Brier 6.9$\times$ and ECE 3.6$\times$ worse than the best graph model. It loses 31\% of its attainable CSI when the validation-frozen threshold is transferred, whereas the GINE variant loses 2\%, and it carries about 2.4$\times$ the seed-to-seed standard deviation. Figure~\ref{fig:predmap}b shows this on one snapshot, where the CNN field is saturated and the GINE field is graded. We draw two conclusions from this comparison. The first is that the training objective explains most of the gap, although the two arms also differ in the loss selected on validation (Section~\ref{sec:models}), so we do not read the objective as the only difference between them. The second is that at matched objective, pooled ranking does not discriminate the formulations, but calibration, threshold transfer and stability favor the graph formulation, which also offers edge-level intervention affordances. %
\subsection{Skill varies with fire characteristics, and pooled scores hide the structure} \label{sec:results-structure} %

This subsection answers RQ2, i.e., where \FGN{}'s skill comes from, and so when the ranked map is most reliable. %

\textbf{Skill falls with track age, for the null as well as for every variant} (Extended Data Table~\ref{tab:edage}), with snapshot-level AP at $\rho = -0.41$ to $-0.47$ across the variants and the steepest fall for the ring-distance null ($\rho = -0.54$). Model-over-null lift persists in every age band for the GINE variant, although it thins to parity at 240--480\,h (1.95$\times$ at 0--48\,h, 1.02$\times$ at 240--480\,h, 1.46$\times$ at 480\,h+).  %

\begin{figure}[!tp] \centering \includegraphics[width=\linewidth]{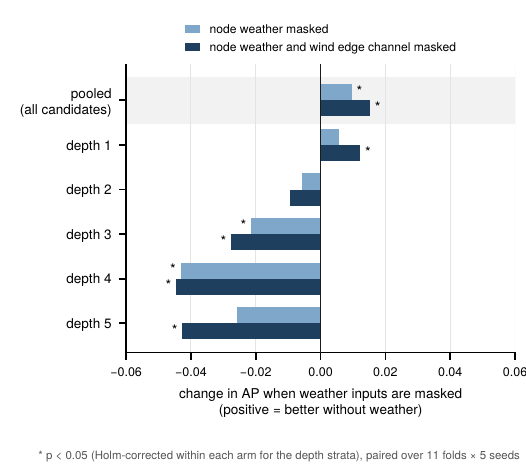} \caption{The weather ablation read two ways. Bars: change in per-snapshot AP when the HRRR inputs are masked (Methods), for the arm that masks the node weather and the arm that also masks the wind edge channel, paired over 11 leave-one-fire-out folds and 5 seeds (Extended Data Table~\ref{tab:edweather}). Pooled over all depths, masking improves AP. Split by ring depth, the gain sits at depth 1 and the losses at depths 3 to 5, where the wind edge channel carries the deep-advance signal. Stars: $p < 0.05$, Holm-corrected within each arm for the depth strata.} \label{fig:pooleddepth} \end{figure}

\textbf{The weather inputs help at depth and hurt at the front, so the pooled change has the opposite sign to the deep-ring change} (Extended Data Table~\ref{tab:edweather}; Figure~\ref{fig:pooleddepth}). Masking all inputs from the High-Resolution Rapid Refresh (HRRR) weather analysis \emph{improves} pooled per-snapshot AP ($+0.0152$, $p = 2.7\times10^{-5}$), because ranking the cells adjacent to the front treats weather as noise. Yet depths 3--5 degrade significantly and near-monotonically without weather, and 10 of 11 development-pool fires get worse (Saddleridge $-0.159$, Dolan $-0.140$). Furthermore, the arm that also masks the wind edge channel moves further than the node-weather arm, both in its gain at depth 1 and in its losses at depths 3--5, so the directional edge features carry signal for deep advance. The cells adjacent to the front are where the distance prior already does most of the work. The deeper rings are where a run outpaces that prior and a model can add information beyond it, so the loss at depth weighs more than the pooled gain suggests. %

\subsection{Enrichments that win on pooled scores do not hold across fires} \label{sec:results-negatives} %

This subsection covers RQ4, which asks whether enrichments, i.e., further inputs or training signals, improve skill. We pre-registered three such experiments, each paired against the same GINE variant, and all three are negative at the fire level (Extended Data Table~\ref{tab:ednegatives}; Supplementary Note~\ref{supp:enrich}). Conditioning on a driver-type label (wind- or fuel-dominated, computed from the weather at $t$) gains $+0.0032$ AP pooled but improves only 5 of 11 fires, and worsens the wind class it was built for ($-0.0143$). An auxiliary fire-intensity head changes AP by $-0.0020$ ($p = 0.71$). A 24-hour auxiliary head gains $+0.0059$ pooled, but 7 of 11 fires favor the 12-hour-only arm. Hence, two of the three would read as wins under pooled-only reporting, which is the pattern the weather ablation also shows. %
\subsection{Model-guided interventions reduce predicted spread most} \label{sec:results-interventions} %

We also show how the graph formulation supports an exploratory analysis of barrier and fuel-treatment interventions on the trained models. We edit the fuel or barrier inputs of $k$ cells and re-score the unchanged model, which is a counterfactual rather than a simulation of fire physics. On fires held out of their leave-one-fire-out fold models (Camp, Kincade and CZU), the reduction in predicted spread grows with the budget $k$, and guided placement reduces it 4--41$\times$ more than random placement at the 25-cell budget on the wind-driven fires and beats a distance-to-front heuristic in all 6 fire$\times$intervention-type combinations (Extended Data Figure~\ref{fig:edinterventions}; Supplementary Table~\ref{tab:interventions}). Whether wind-aligned barrier sets suppress predicted spread more than crosswind ones differs between fires. These are sensitivities of a learned model rather than estimates of suppression effect, and on the fuel-dominated CZU fire random barrier edits even increase predicted spread (Supplementary Note~\ref{supp:interventions}). %

\section{Discussion} \label{sec:discussion}
The main finding is that the directed graph formulation provides measurable skill in ranking near-front growth beyond proximity alone, while the apparent importance of particular modeling choices depends on where and how that skill is measured. The \FGN{} variants achieve average precision of 0.0624--0.0710 against 0.0369 for the distance baseline on the California test set (Table~\ref{tab:main}), and the GINE variant achieves 0.0964 against 0.0595 on the out-of-state set (Table~\ref{tab:w2}). Weather masking reveals a spatially differentiated signal: the pooled statistic improves, but performance deteriorates farther from the front and on most development fires (Figure~\ref{fig:pooleddepth}; Extended Data Table~\ref{tab:edweather}). This result supports a role for environmental information in deeper advance that is obscured by aggregation over the more numerous near-front cases. Under the tested settings, the GINE variant has lower raw miscalibration and more stable threshold behavior than the objective-matched CNN%
. The enrichment experiments likewise show that a favorable snapshot-level statistic need not translate into improvement across fires (Extended Data Table~\ref{tab:ednegatives}). The contribution is thus a spatially explicit account of predictive skill and its measurement, together with an editable representation supporting intervention responses.%

The architectural contribution of this paper is the formulation itself. The evidence is that neighborhood context around the front is necessary and the edge features carry signal, but the choice of aggregator does not measurably change ranking (Supplementary Note~\ref{app:mechanism}). %
 Against the raster alternative, the combined evidence supports the graph formulation as the better-suited approach, since the best-calibrated variant, GINE, matches the objective-matched CNN on pooled ranking and does better on calibration, threshold transfer and seed stability. %

Four of our own results show that the score a spread model reports depends on how the evaluation is built and read. The first two come from how the evaluation is constructed. Under the hindsight candidate set of our first protocol, a feature-free depth ranking recovers $\sim$88\% of the best model's AP and deep-first ranking looks informative (Figure~\ref{fig:candsets} and Supplementary Note~\ref{sec:v1}). The final-perimeter-conditioned evaluation of \citet{roesch2024} shares this hindsight construction in a different form. On the prediction-time candidate set, however, the same deep-first ranking is anti-predictive (below prevalence; Table~\ref{tab:main}). Under a footprint training objective, the same CNN sits below the distance null and recovers to a level statistically indistinguishable from the graph models once re-supervised on the candidate objective. The other two come from pooled-only reporting. The 24\,h auxiliary head and the driver-conditioning feature variant are statistically significant pooled wins that reverse at the fire level (Extended Data Table~\ref{tab:ednegatives}). Similarly, the weather ablation reverses between its pooled and its stratified reading (Section~\ref{sec:results-structure}). Hence, the practical prescriptions are the ones this paper's protocol enforces, namely deterministic nulls beside the headline and beside each stratified result, strata next to pooled statistics, and fire-level cross-checks on paired claims. %

\phantomsection\label{sec:limitations}The findings describe retrospective, 12-hour growth prediction within five H3 rings of an active front; this region contains 88.8\% of newly burned cells in the California test set but omits substantial growth in some rapid runs, motivating wider candidate regions and models with appropriate spatial reach (Supplementary Note~\ref{app:coverage}). Satellite-derived labels introduce detection uncertainty and variable observation intervals, while the approximately 8.5-hour offset between nominal FEDS timestamps and observation times requires explicit treatment of observation, data availability, and forecast issuance in prospective evaluation (Section~\ref{sec:taskdef}). Near-kilometer cells, historical fuel layers, and weather sampled before the growth interval limit representation of fine-scale barriers and changing conditions, motivating refreshed fuels, temporally aligned weather, and higher-resolution tests (Section~\ref{sec:features}). The small number of independent test fires, disclosed repeated in-state test contacts, and the modest selection budgets on both sides restrict comparative inference; further studies should use additional untouched fire seasons, larger search budgets, and uncertainty estimates that account for dependence within fires (Supplementary Notes~\ref{app:contacts} and~\ref{app:tuning}). Furthermore, the probabilities remain uncalibrated, and their use requires calibration on separate data and evaluation under deployment-specific thresholds and costs. The value of the enrichments at other capacities, resolutions, or supervision scales remains unresolved. Finally, fuel and barrier edits measure sensitivities of an observationally trained model, including nonphysical responses in some cases, so intervention-aware learning and evaluation against independent simulations or documented treatment outcomes are needed before claiming management effectiveness. A staged pathway from retrospective replay to shadow deployment is set out in Supplementary Note~\ref{supp:pathway}.

\phantomsection\label{sec:conclusion}

The practical contribution is a representation that can translate an observed fire perimeter and environmental inputs into a ranked map of locations susceptible to near-front growth during the next observation interval. Such rankings could help fire analysts focus subsequent assessment on particular flanks and relate potential growth to exposed communities, access routes, and response resources. Building-scale damage prediction \citep{esparza2026graphfire} and post-fire damage assessment from ground-level imagery \citep{esparza2025vlmdamage} address the later end of the same chain, from exposure to loss. The out-of-state results provide initial evidence for geographic transfer, while variation among fires and incomplete coverage of rapid runs define conditions requiring local validation. The same representation allows fuel and barrier attributes to be edited, so that the model's responses to candidate interventions can nominate scenarios for independent physical simulation and expert review, although they do not estimate the realized effectiveness of suppression or fuel treatment. Operational adoption therefore requires timely data delivery, probability calibration, prospective testing, and independent validation of intervention responses. %
\section{Methods} \label{sec:data} \label{sec:task}
 Two design choices run through every stage of the pipeline, and we state them here since the rest of the section applies them repeatedly. The first is that no input and no candidate cell may use anything that is not known at prediction time. The second is that every score is placed beside a simple reference rule and read per fire and per stratum, not only pooled over all cells.  Figure~\ref{fig:formulation}a gives an overview of how one snapshot is built, from the FEDS perimeters to the per-cell and per-edge inputs that the models consume. %
\subsection{Task definition} \label{sec:taskdef}
 The labels and the fire geometry come from FEDS, the Fire Event Data Suite \citep{chen2022feds}, which assembles Visible Infrared Imaging Radiometer Suite (VIIRS) 375\,m active-fire detections into object-tracked fire perimeters at 12-hour cadence. We use its public western-US deposit covering the 2012--2021 seasons \citep{feds_zenodo}. Each fire in our data is a sequence of perimeter snapshots at times $t_0, t_1, \ldots$ that are about 12 hours apart (Extended Data Figure~\ref{fig:edsubdaily}). These stamps are half-day labels rather than observation times. Detections from the daytime VIIRS overpass (about 13:30 local time) form the perimeter stamped 12:00 UTC, and those from the night overpass (about 01:30 local time) form the perimeter stamped 00:00 UTC of the following day. A perimeter stamped $t$ is therefore observed about 8.5 hours after $t$, and the growth we label between $t$ and $t + 12$\,h occurs between the two overpasses, roughly from $t + 8.5$\,h to $t + 20.5$\,h. The window stamped 12:00 UTC spans the afternoon-to-night burning period and the window stamped 00:00 UTC the night-to-morning period. The candidate sets and labels are defined from one stamp to the next and are unaffected by this offset.

At each snapshot time $t$ we build one graph. The nodes are H3 resolution-8 hexagonal cells, which are about 0.74\,km$^2$ each with about 0.9\,km between neighboring cell centers \citep{h3uber}.\footnote{\url{https://h3geo.org}.}  The node set at time $t$ consists of the burned footprint plus the unburned cells within five rings of the active front, where the active front is the set of burned cells that border at least one unburned cell (the choice of five rings is analyzed in Supplementary Note~\ref{app:coverage}). Neighboring cells are connected by two directed edges, one in each direction. We use a directed graph because the quantities that govern spread depend on direction, in that wind pushes fire from one cell into another and fire climbs uphill differently than it descends. Each directed edge therefore carries its own five features (Section~\ref{sec:features}), and rather than having fixed edge weights, the message-passing layers learn how much to weight each edge from its features. A typical snapshot graph has about 2{,}750 nodes and 16{,}000 directed edges (median: range 102--9{,}316 nodes across the 909 graphs). 

The task itself is node classification on a prediction-time-constructible candidate set. Specifically, the candidates are the unburned cells within 5 H3 rings ($\sim$4.5\,km) of the \emph{current} front, which comes to a median of about 1{,}250 cells per snapshot. The label $y \in \{0, 1\}$ marks whether the candidate cell is inside the fire perimeter at $t + 12$\,h, i.e., whether it newly burns during the next 12-hour window. Already-burned cells serve as context for message passing and are masked with $y = -1$ from both the loss and every metric. In other words, the labels are the observed outcome of the coming window, which is known in hindsight for training and evaluation, and no input feature is computed from it. Historic burns enter only as input features (burn history; Section~\ref{sec:features}). Positive labels are rare, at 22{,}247 of 1{,}184{,}772 candidate cells corpus-wide (1.9\%) and 1.65\% on the test pool. 

 The hindsight candidate construction we first used, and why it was replaced, is analyzed in Supplementary Note~\ref{sec:v1}, and Figure~\ref{fig:candsets} shows the two constructions on the same snapshot. %

\subsection{Scope of the claims} \label{sec:scope}
All claims in this paper are scoped in four ways. The first is time and place, in that they cover 2018--2021 California fires plus the 2020--2021 out-of-state confirmation set, because the public FEDS archive available when the corpus was frozen ended with the 2021 season \citep{feds_zenodo}. A 2012--2024 FEDS-MTBS release (MTBS, Monitoring Trends in Burn Severity), used by \citet{kondur2026pyrostack}, has since appeared. The second is near-front spread, i.e., candidate cells within 5 H3 rings of the active front, which holds 85--89\% of the realized 12\,h growth in state and 72\% on the out-of-state set (58--94\% per fire) (Section~\ref{sec:corpus} and Supplementary Note~\ref{app:coverage}). The third is actively growing fires, since snapshots pass a backward-looking activity gate, i.e., growth in the prior 24\,h or a track younger than 24\,h. So our claims concern where an active fire spreads next rather than whether a dormant fire resumes growth. Lastly, they cover the 12\,h horizon set by the FEDS cadence. All reported probabilities are raw model outputs, and no post-hoc calibration is applied anywhere. %
\subsection{Features} \label{sec:features}
Each node carries 81 stored features plus a half-day flag, appended at load time, which records whether the snapshot carries the 12:00 or the 00:00 UTC stamp. These come from 6 public sources, and all of them are estimated at the cell and computable at prediction time, as listed below.
\begin{itemize}
\item \textbf{Landscape} (64 dims): the AlphaEarth Foundations annual satellite embedding \citep{alphaearth2025, aef_catalog}, obtained from Google Earth Engine,\footnote{Earth Engine collection \nolinkurl{GOOGLE/SATELLITE_EMBEDDING/V1/ANNUAL}, \url{https://developers.google.com/earth-engine/datasets/catalog/GOOGLE_SATELLITE_EMBEDDING_V1_ANNUAL}.} using the version for the year before the fire. The same embeddings enrich the node features of a recent wildfire building-damage GNN \citep{esparza2026graphfire}. %
\item \textbf{Terrain} (4 dims): elevation at the cell centroid from the USGS 3D Elevation Program \mbox{\citep{usgs3dep}},\footnote{\url{https://www.usgs.gov/3d-elevation-program}; point queries through the Elevation Point Query Service (\url{https://epqs.nationalmap.gov}) for the in-state corpus, bulk sampling from the 3DEP image service for the out-of-state set.} with slope and aspect (as sine and cosine) derived numerically from neighboring centroids. Because the centroids are about 0.9--1.0\,km apart, slope is a mesoscale quantity over a window of roughly 1.9\,km and understates site slope.
\item \textbf{Fire history} (2 dims): the number of prior fires that burned the cell and the days from the last prior burn to the fire's start, computed from the FEDS archive itself\footnote{FEDS western-US deposit, \url{https://doi.org/10.5281/zenodo.15261836}.} with the current fire excluded.
\item \textbf{Weather} (5 dims): NOAA HRRR 3\,km analyses \citep{dowell2022hrrr, hrrr_aws} valid at $t$,\footnote{Retrieved from the NOAA HRRR archive on AWS Open Data, \url{https://registry.opendata.aws/noaa-hrrr-pds/}.} 2\,m temperature, relative humidity, 10\,m wind speed, boundary-layer height, and a precipitation field that is identically zero (below).
\item \textbf{Fuels} (4 dims): the LANDFIRE LF2016 products \citep{landfire}:\footnote{\url{https://landfire.gov}; sampled from the LANDFIRE Product Service image servers.} the Scott and Burgan 40 fuel model \citep{scottburgan2005}, embedded as a categorical variable, and canopy cover, height and base height. Every corpus fire draws the LF2016 products because LF2020, the first version to map the 2017--2020 fuel disturbance, was released in 2022, after every corpus fire, and the interim LF2019L update covered submitted events only. Recent burn scars are therefore invisible to the fuel layers for the 2020--2021 fires, which the FEDS-derived burn-history features partly compensate.
\item \textbf{Flags} (2 stored dims + the half-day flag): whether the cell is on the active front, and whether it lies within one ring of the fire's recorded ignition point (ignition points from GOFER, the GOES-Observed Fire Event Representation \citep{liu2024gofer}, where available, otherwise the reported ignition coordinates; for 4 fires with neither, the centroid of the first FEDS snapshot). %
\end{itemize}
Each directed edge from cell $a$ to cell $b$ carries 5 features, as listed below.
\begin{itemize}
\item the compass bearing of the edge;
\item the cosine of the angle between that bearing and the HRRR wind direction at cell $a$, so that an edge pointing downwind scores $+1$ and one pointing into the wind $-1$;
\item the elevation difference $b - a$, positive when the edge climbs;
\item a fuel-continuity score in $\{0, 0.5, 1\}$ from the two cells' fuel models (1 if both are burnable and in the same fuel category, 0.5 if burnable but different or if either is unknown, 0 if either is a hard break);
\item a barrier indicator that is 1 when either cell is non-burnable (water, snow/ice, urban, barren or agriculture).
\end{itemize}
 Every feature respects the prediction-time rule: landscape embeddings use the year before the fire, LANDFIRE versions are strictly before the fire year, burn history excludes the current fire, and the standardizer is fit on training fires only.

Weather features are taken from the HRRR analysis valid at the stamp time $t$, using the most recent model cycle started at or before $t$.  This keeps the inputs strictly prediction-time valid, since a response team holds this analysis well before the perimeter is available and forecast skill cannot leak into the labels. It does, however, understate the weather signal, and forecast weather aligned to the observation interval is left to future work. Precipitation is the one weather variable that carries no information in our features. Specifically, HRRR stores precipitation as an amount accumulated over the forecast interval, and at analysis time that interval has zero length, so the stored field is zero everywhere. A usable precipitation feature would need the one-hour accumulation from the first forecast step, which we leave for future corpus work. 

\subsection{Corpus and splits} \label{sec:corpus}
 The corpus contains 16 fires with 909 snapshot graphs (California, 2018--2021). The in-state split assigns 8 fires to training (Camp 2018, Carr--Hirz--Delta 2018, Mendocino Complex 2018, Woolsey 2018, Kincade 2019, Saddleridge 2019, CZU Complex 2020, August Complex 2020), 2 to validation (North Complex 2020, Dolan 2020) and 5 to the sealed test (Creek 2020, Bobcat 2020, Glass 2020, Dixie 2021, Caldor 2021), and Supplementary Table~\ref{tab:corpus} lists the role of every fire. The split is fire-disjoint and temporally ordered, in that every training and validation fire ignited before every test fire, so there is no temporal leakage across the split. While fire tracks overlap in calendar time, no snapshot of any fire appears on both sides of the split. The 16th fire, Hennessey (LNU) 2020, is held out of this split. Together with the 10 training and validation fires it forms an 11-fire development pool, which we use for every paired leave-one-fire-out experiment (the ablations and expansions of Section~\ref{sec:results}) and, with North Complex and Dolan as the validation pair, to train the model evaluated out of state. The sealed test pool has 370 graphs, 512{,}742 candidate cells, and 8{,}455 positives (prevalence 0.0165). Snapshots pass the backward-looking activity gate of Section~\ref{sec:scope}, and 120 zero-growth windows are retained (48 in the test pool), which we verified as true zero-growth over all 16 fires (the FEDS tracks for Bobcat 2020 and Saddleridge 2019 inherit ids from earlier activity, so their pre-ignition snapshots survive only as zero-growth windows, and Supplementary Table~\ref{tab:corpus} reports activity-gated snapshot counts rather than raw track spans, which overstate active duration because tracks persist through dormancy). %

The out-of-state confirmation set comprises 5 fires from the 2020--2021 seasons, 3 in Oregon (Santiam Complex, Holiday Farm, Bootleg), 1 in Washington (Cold Springs--Pearl Hill) and 1 in Colorado (East Troublesome), amounting to 175 graphs, 194{,}796 candidates and 5{,}952 positives (prevalence 0.0306). It was built with the same pipeline and rules (terrain sampled in bulk from the same 3DEP service, ignition points from the reported coordinates) and SHA256-frozen at build time, and it was never used for training or validation. We defined the transfer claim requirement as a pooled lift of $\gtrsim 1.5\times$ \emph{and} an established-track stratum whose lift also exceeds 1 (mechanics in Supplementary Note~\ref{app:w2}). We note that the 5-ring horizon covers less of the growth out of state, at 71.9\% of newly burned cells pooled. Per fire, it ranges from 58\% on East Troublesome and 59\% on Santiam Complex to 64\% on Cold Springs--Pearl Hill, 73\% on Holiday Farm and 94\% on Bootleg. %

Two notes on the corpus follow from the 5-ring horizon. The first is that the horizon does not capture all growth, since some cells that burn within a 12-hour window lie more than 5 rings ahead of the front, mostly during long wind-driven runs or spotting, and these cells are not candidates and so cannot be scored. The cov@5 column of Supplementary Table~\ref{tab:corpus} reports, for each fire, the fraction of newly burned cells that fell within 5 rings. For Woolsey it is 63\%, meaning that 37\% of Woolsey's new-burn cells were beyond the horizon, while for Saddleridge it is 100\%.  On the test pool it is 88.8\% overall, with Glass lowest at 58\%, where most of the missing growth is in its first window because the fire was already large at first detection and the initial front was far behind the actual perimeter. We note that the 2 training fires labeled wind-driven grass/shrub (Woolsey and Kincade) are among the lowest-coverage fires, and this matters for how our results should be read. Long wind-driven runs are the cases where wind direction and slope should help a model the most, yet those are also the cells most often left out by the 5-ring horizon.  The second note is that the first snapshot of a fire track (track age below 12\,h) behaves differently from later ones because the first perimeter of a track lags the fire\textquoteright s true extent (detection latency, as in the Glass example above), so in the out-of-state table we report it as its own stratum, while in-state it falls inside the 0--48\,h track-age band. Supplementary Note~\ref{app:coverage} shows why widening the horizon does not help (coverage by ring in Supplementary Table~\ref{tab:ringR}). %

One further note concerns how the test pool is weighted, in that the 2 largest test fires, Creek and Dixie, burned for months, so most of their snapshots come late in the fire. We measure this by track age, which is the time since the fire was first detected. Snapshots more than 480\,h (20 days) into a fire hold 78\% of all test candidate cells but only 52\% of the positives, so pooled test metrics are dominated by late, slow-growth snapshots. Hence, we report results split into track-age bands alongside the pooled numbers (Extended Data Table~\ref{tab:edage}), where the label is the same in every band, i.e., newly burned cells in the next 12 hours, and only the snapshots being scored differ.

The sealed in-state test was scored five times under this protocol, each pass declared or disclosed (Supplementary Note~\ref{app:contacts} lists every contact). No selection or code change was conditioned on a test read, and the one decision that followed a test read, namely taking the objective-matched raster arm to the test, is disclosed there as well. The out-of-state set was evaluated exactly once under pre-registration. The age bands re-score the saved sealed-test predictions, whereas the weather ablation is a paired development-pool experiment with newly trained models and no test contact. It masks the HRRR inputs by setting the node weather to its training mean and the wind alignment to zero. %

\subsection{Model families} \label{sec:models}
\FGN{} is the formulation of Section~\ref{sec:taskdef} together with a message-passing model on it, and we compare 5 message-passing architectures and one control within it, all trained on the same directed graphs and implemented in PyTorch Geometric \citep{fey2019pyg}. Where a single model is needed, as in the out-of-state pass, the intervention probe and the GINE columns of Figure~\ref{fig:predmap}, it is the GINE variant, which the results select as the best calibrated of the 5. Figure~\ref{fig:formulation}b shows the architecture they share and the point at which each one admits the edge features. This set is chosen to answer RQ1, since each model admits the directional edge information at a different point in the computation, from not at all to both the message and the attention weight (Figure~\ref{fig:formulation}b). Through this comparison we are able to tell whether message passing is needed at all, whether the edge features help, and whether the way they are combined (the aggregation mechanism) matters. %
\begin{itemize}
\item \textbf{MLP (control).} Scores each cell from its own features alone. Since there is no message passing, a cell knows nothing about its neighbors or the front, which makes this the baseline that every variant must beat for message passing to be considered necessary.
\item \textbf{GraphSAGE} \citep{hamilton2017} (SAGE in the tables and figures). Each cell updates its representation with the mean of its neighbors' representations. The edge features are ignored, so this model is edge-blind, and it tests how much skill comes from neighborhood context alone without any directional information.
\item \textbf{SAGE-maxpool.} Uses the same edge-blind GraphSAGE message passing, but computes the output differently. Each directed edge into a cell is scored from the two endpoint representations and the edge features, and the cell's score is the maximum over its incoming edges, which treats spread as the single most likely path into a cell. Hence, it tests whether directional information is useful when it enters only at the output.
\item \textbf{GINE} \citep{xu2019gin, hu2020gine}. Each cell sums the messages from its neighbors, where each message is adjusted by adding a transformation of the edge features before the sum. So wind alignment, slope and fuel continuity change the content of every message, which is the simplest way to put directional information inside the message passing itself.
\item \textbf{GATv2} \citep{brody2022gatv2, velickovic2018gat} (graph attention network, version 2). Each cell attends to its neighbors with learned weights, and the edge features enter the computation of those weights. So the model learns which neighbors to listen to based on the direction of wind and slope along the edge, while the edge features do not change the content of the messages, only their weights.
\item \textbf{Graph Transformer} \citep{shi2021unimp}. Also attention based, but here the edge features enter both the attention weights and the message content. This is the strongest pathway for directional information among the 5, and it tests whether a richer use of the edge features improves on GINE and GATv2.
\end{itemize}
All are trained with focal loss \citep{lin2017focal} on candidate cells only (Section~\ref{sec:taskdef}).

The raster baseline, against which the graph formulation is compared in Section~\ref{sec:results-raster}, is a Res18-U-Net \citep{ronneberger2015unet, he2016resnet, iakubovskii2019smp}, which is the WildfireSpreadTS-standard architecture. The inputs were the same node features rasterized to the H3 pitch plus wind $u/v$ channels (a CNN is otherwise direction-blind) and a valid-cell mask, trained in 2 arms. One arm is trained on the burned footprint (full-mask segmentation of every cell burned by $t+12$\,h), and the other is an objective-matched arm supervised on exactly the candidate cells the graph models use. Each arm took the best of its own 6-configuration validation sweep over learning rate and loss, which selected focal loss for the footprint arm and weighted binary cross-entropy for the objective-matched arm. %

Trained on the burned footprint, the CNN ranks 12\,h spread below the distance null (Section~\ref{sec:results-raster}). This is because the segmentation target is about 59\% positive once the burned interior is included, so continued training on it degrades the ranking at the front. In fact, every seed reached its best validation epoch within 8 epochs, and predictions overshoot prevalence about fifteen-fold on average. The out-of-state result is measured for the graph model only (Section~\ref{sec:results-w2}). We expect the CNN to transfer worse, since the properties that transfer stresses, namely threshold transfer, calibration and stability, are the ones on which it already loses in-state. An objective-matched CNN pass on the out-of-state set would settle this, although it would be a second, declared contact with that set and has not been run.

Hyperparameters for every variant were selected by leave-one-fire-out search over development fires only, and final models were retrained on 10 fresh seeds with validation-frozen thresholds (selection procedure and budgets in Supplementary Note~\ref{app:tuning}). The MLP control was deliberately left untuned and trained at the repository defaults.

\subsection{Null models} \label{sec:nulls}
 The headline table carries 3 of the 5 deterministic null models described below (persistence and uniform buffer score exactly the prevalence and are not tabulated), which need no training and have no seed variance, and hence appear as table rows and never inside the seed-paired significance tests. For PR-AUC, no skill equals the prevalence (0.0165 on the test pool).
\begin{itemize}
\item \textbf{Ring distance, near-first.} Scores each candidate by how close it is to the front, with ring 1 highest and ring 5 lowest. This captures the prior that fire spreads to adjacent cells first, and since it is the strongest null, a model must beat it to show any skill beyond distance.
\item \textbf{Wind--slope heuristic.} A fixed rule on the edge quantities the models see, with signs taken from the Rothermel spread model \citep{rothermel1972}, in which fire spreads faster with the wind and uphill, needs continuous fuel, and is stopped by non-burnable ground. Each directed edge gets a score of $1.5 \times$ wind alignment $+$ a bounded uphill term $+$ fuel continuity $- 3 \times$ barrier, squashed to a probability, and each cell takes the maximum over its incoming edges. The magnitudes were set by hand before any experiment and never tuned, so this null is a floor rather than a calibrated physical model.
\item \textbf{Persistence.} Predicts no new spread anywhere. Under a target that includes fire already present at prediction time (persisting active fire in NDWS and WildfireSpreadTS) this rule scores highly, reaching 0.115 against a best of 0.284 in NDWS and 0.193 against 0.372 in WildfireSpreadTS \citep{huot2022, gerard2023}. Under our new-growth target, in contrast, it has no skill by construction and its PR-AUC equals the prevalence. %
\item \textbf{Uniform buffer.} Predicts that every candidate cell burns, which is the ``evacuate the whole buffer'' rule, and its PR-AUC also equals the prevalence.
\item \textbf{Ring distance, deep-first.} The reverse of the first null, with ring 5 highest and ring 1 lowest. This is an anti-null, in that it scored well under our first protocol because that candidate set was built from where the fire went (Supplementary Note~\ref{sec:v1}), whereas under a candidate set built at prediction time it should score below chance. Its below-prevalence PR-AUC (0.0125, Table~\ref{tab:main}) and below-chance area under the receiver operating characteristic curve (ROC-AUC) of 0.26 confirm that the test candidate set carries no hindsight. %
\end{itemize}
The strongest null is carried into every stratified table (per fire, per track-age band, and out of state). The paired treatment-versus-control experiments need no null row, because both arms face the identical candidate population.

\subsection{Metrics and statistical tests} \label{sec:metrics}
The primary score is average precision (AP, also reported as PR-AUC), which is the area under the precision--recall curve over the candidate cells. It measures how well the model ranks the cells that burned above those that did not, and it is the right score for a rare event because it ignores the many easy negatives. A model with no skill scores the prevalence, i.e., the share of candidates that burned. So we report every AP next to the prevalence and next to the strongest null model, and we call the ratio of the model's AP to the null's AP the lift. A pooled score is computed over all candidate cells at once, whereas a per-fire (macro) score is averaged over fires so that every fire counts equally. A stratum is a subgroup of snapshots or cells, such as one ring depth or one band of track age. We also give ROC curves (Extended Data Figure~\ref{fig:edsealed}e) so that our models can be read against studies that quote ROC-AUC, although it is less informative than AP when positives are rare. Since a warning has to be issued at a fixed threshold, we also report the critical success index (CSI), which is the hit rate at a decision threshold $\tau$ once false alarms are counted against it. We give it both at the best possible threshold (CSI$_{\max}$) and at the threshold frozen on the validation fires (CSI@$\tau$), which is the operational case. Calibration asks whether a predicted probability of 0.3 burns about 30\% of the time. We report it through the Brier score, which is the mean squared error of the probabilities, and the expected calibration error (ECE), which is the mean gap between predicted and observed frequency over 10 equal-width probability bins. Reliability diagrams show the same comparison bin by bin (per-architecture values in Supplementary Table~\ref{tab:calibration}). %

The statistical tests are chosen to follow how the data are paired, since seeds, snapshots and fires each give a different unit of comparison. Across the 6 trained models we use a Friedman test on the per-seed pooled AP and, for each model against the best-ranked one, a paired Wilcoxon signed-rank test over the 10 seeds with Holm correction for the multiple contrasts. All tests are two-sided. The raster arms were trained with their own seeds, so the CNN-versus-GNN contrast uses a Welch two-sample $t$-test. For every paired treatment-versus-control experiment we report two statistics. The pooled per-snapshot (micro) statistic is a paired Wilcoxon test over the seed-averaged AP of every snapshot. The fire-level (macro) statistic is the mean per-fire change, with the count of fires that improved and a one-sample Wilcoxon test over the 11 per-fire changes. We report both because the micro test alone can produce significance from a few large fires, which is a pattern we take up in Section~\ref{sec:discussion}. Correlations with track age use Spearman's $\rho$. Average precision is computed with a tie-aware implementation, where tied scores are scored as a block so that the statistic is permutation-invariant, since an order-dependent AP overstated a tied depth scorer's 0.306 as 0.356 in one of our experiments (Supplementary Note~\ref{sec:v1}, where the best model's AP was 0.346).

The Friedman test across the 6 trained models is significant ($\chi^2 = 26.6$, $p = 10^{-4}$), but only because of the control (Holm $p = 0.010$). We read the tie among the variants against a power analysis. With 5 test fires the smallest paired difference detectable at 80\% power is 0.015--0.036 AP depending on the pair, and 0.032--0.053 AP on the 11-fire development pool, whose fires differ more from one another. The observed differences between the variants are 0.003--0.009 AP, so deciding among the aggregators would need about 10 to 40 held-out fires for a 0.01 AP difference and 30 to 150 for a 0.005 difference. The model-versus-null and model-versus-control contrasts are decidable on the 11-fire development pool. On the 5 sealed fires the GINE-minus-null difference (0.054) is below the minimum detectable effect at that size (0.072), although GINE exceeds the null on every sealed fire (Supplementary Table~\ref{tab:edperfire}).

\subsection{Use of AI assistance} \label{sec:ai}
We used an AI coding assistant (Claude, Anthropic) under the authors' direction for editing code and figure scripts, for editing manuscript text, and for checking the reported numbers against the saved analysis records and the cited literature. All outputs were reviewed and verified by the authors, who take full responsibility for the content, and the assistant is not an author of this work.
\section*{Data availability}
All input data used in this study are publicly available from the following sources.
\begin{itemize}
\item The fire perimeters, burn history and fire radiative power come from the Fire Event Data Suite (FEDS) western-US archive, deposited at Zenodo (\url{https://doi.org/10.5281/zenodo.15261836}) \citep{feds_zenodo}, following the FEDS method \citep{chen2022feds} (\url{https://doi.org/10.1038/s41597-022-01343-0}).
\item The weather analyses come from the NOAA High-Resolution Rapid Refresh (HRRR) archive on the AWS Open Data Registry (\url{https://registry.opendata.aws/noaa-hrrr-pds/}) \citep{dowell2022hrrr, hrrr_aws}.
\item Fuel models and canopy layers come from the LANDFIRE LF2016 products (\url{https://www.landfire.gov/data/lf2016}), which we sampled through the LANDFIRE Product Service (\url{https://lfps.usgs.gov/arcgis/rest/services}) \citep{landfire}.
\item Elevation comes from the USGS 3D Elevation Program (\url{https://www.usgs.gov/3d-elevation-program}). Specifically, we queried it through the Elevation Point Query Service (\url{https://epqs.nationalmap.gov}) for the California corpus and through the 3DEP image service (\url{https://elevation.nationalmap.gov/arcgis/rest/services/3DEPElevation/ImageServer}) for the out-of-state set \citep{usgs3dep}.
\item The landscape embeddings are the AlphaEarth Foundations annual satellite embeddings, available as the Google Earth Engine collection \nolinkurl{GOOGLE/SATELLITE_EMBEDDING/V1/ANNUAL} (\url{https://developers.google.com/earth-engine/datasets/catalog/GOOGLE_SATELLITE_EMBEDDING_V1_ANNUAL}) \citep{alphaearth2025, aef_catalog}.
\item Ignition points come from the GOES-Observed Fire Event Representation (GOFER) product \citep{liu2024gofer} (\url{https://doi.org/10.5194/essd-16-1395-2024}; data \url{https://doi.org/10.5281/zenodo.8327264}) where it covers a fire. Otherwise they come from reported ignition coordinates in the Fire Program Analysis fire-occurrence database (sixth edition, 1992--2020; \url{https://doi.org/10.2737/RDS-2013-0009.6}) \citep{short2022fpafod} and in public incident records.
\item The pressure-level fields behind the driver-type label are ERA5 hourly data on pressure levels from the Copernicus Climate Data Store (\url{https://doi.org/10.24381/cds.bd0915c6}) \citep{hersbach2020era5}.
\item Cells are indexed on the H3 hexagonal grid (\url{https://h3geo.org}) \citep{h3uber}.
\end{itemize}
The derived data are deposited together with the code snapshot in a restricted-access Zenodo record (DOI 10.5281/zenodo.23272584), which is shared with the editors and reviewers through a secret link during review. The record will be released publicly at publication under the CC BY-NC 4.0 license that the FEDS source data carry. Specifically, the deposit holds the 16-fire California graph corpus and the out-of-state graph corpus, which was frozen with SHA256 checksums at build time. Alongside the graphs it holds the driver-type label table, computed from information available at time $t$, and the fire radiative power (FRP) target table. It also holds the out-of-state pre-registration document and the saved predictions and analysis records from which every table and figure in this paper regenerates without retraining, with every file listed beside its checksum. We note that publishing the out-of-state labels ends their sealed status, so later work should treat that set as held out rather than sealed.

\section*{Code availability}
The code used in this study is available to the editors and reviewers through the same restricted-access Zenodo record. It will be released at publication under the MIT license, with the record DOI as its citable identifier. The repository contains the graph builders, which turn the public inputs above into the per-snapshot graphs, as well as the training and hyperparameter-search code for the graph models and the raster baseline. It also contains the evaluation code with the null models and statistical tests, and the generators that rebuild every table and figure from the saved predictions. A demonstration (\texttt{scripts/demo\_one\_fire.py}) scores one development fire with fold models trained without it and runs on a standard CPU in about 7 seconds. The models were implemented in Python 3.11 with PyTorch Geometric 2.8.0 and PyTorch 2.8.0 (CUDA 12.8) on the Delta system, or PyTorch 2.6.0 (CUDA 12.4) on a lab workstation. Hyperparameters were searched with Optuna 4.5.0 (the selection studies on Delta; 4.9.0 in the lab environment), cells were indexed with the h3 library 4.5.0, and the raster baseline used \texttt{segmentation\_models\_pytorch} 0.5.0. %
\newcommand{\doi}[1]{doi:~\href{https://doi.org/#1}{\nolinkurl{#1}}}
\bibliographystyle{plainnat}
\bibliography{references,references_revision} %
\ifanonymous\else
\section*{Acknowledgements}
This work used the Delta system at the National Center for Supercomputing Applications from the Advanced Cyberinfrastructure Coordination Ecosystem: Services \& Support (ACCESS) program, which is supported by U.S. National Science Foundation grants \#2138259, \#2138286, \#2138307, \#2137603, and \#2138296. We also acknowledge the support of the members in Urban Resilience AI lab at Texas A\&M University.

\section*{Author contributions}
A.P.K., M.E. and A.M. conceived the study. A.P.K. built the graph corpora, designed the prediction task and the evaluation protocol with M.E. and A.M., wrote the software, ran the experiments and the statistical analysis, prepared the figures and wrote the first draft. M.E. and A.P.K. checked the results against the saved artifacts. M.E. contributed to the writing and editing of the manuscript. A.M. supervised the work and acquired the funding. All authors reviewed and edited the manuscript and approved the final version.

\fi

\section*{Competing interests}
The authors declare no competing interests.
\appendix
\clearpage
\expandafter\let\expandafter\origfigstar\csname figure*\endcsname \expandafter\let\expandafter\origendfigstar\csname endfigure*\endcsname
\expandafter\let\expandafter\origtabstar\csname table*\endcsname \expandafter\let\expandafter\origendtabstar\csname endtable*\endcsname
\onecolumn
\renewenvironment{figure*}[1][]{\begin{figure}[H]}{\end{figure}}
\renewenvironment{table*}[1][]{\begin{table}[H]}{\end{table}}
\newcommand{\EDTitle}{Extended Data}
\EDnumbering
\section*{\EDTitle}
The figures and tables in this section are referenced from the main text and the Methods and are collected here in order of first reference.
\begin{figure*}[!tp] \centering \includegraphics[width=\textwidth]{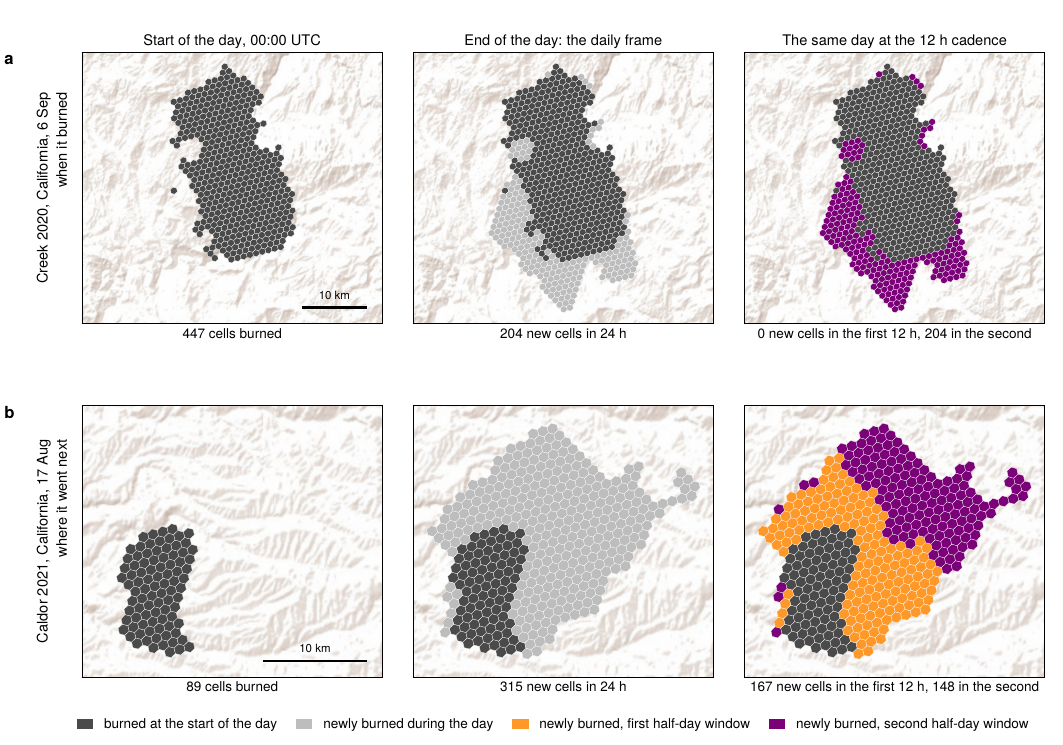} \caption{Two days of observed spread at the half-day FEDS cadence, one perimeter per satellite overpass, against the daily frame of next-day products. (a) Creek (California), 6 September 2020, the day of the snapshot in Figures~\ref{fig:formulation} to~\ref{fig:candsets}: no cells burned in the first 12 hours and 204 in the second, which the daily frame places anywhere within 24 hours. (b) Caldor (California), 17 August 2021: 167 cells in the first 12 hours and 148 in the second, on different flanks, which the daily frame merges with no order of arrival. The FEDS windows follow the VIIRS overpasses, so they are about, not exactly, 12 hours apart. Basemap tiles: Esri World Terrain.} \numlabel{figure}{fig:edsubdaily}\label{fig:subdaily} \end{figure*}

\begin{figure*}[!tp] \centering \includegraphics[width=0.94\textwidth]{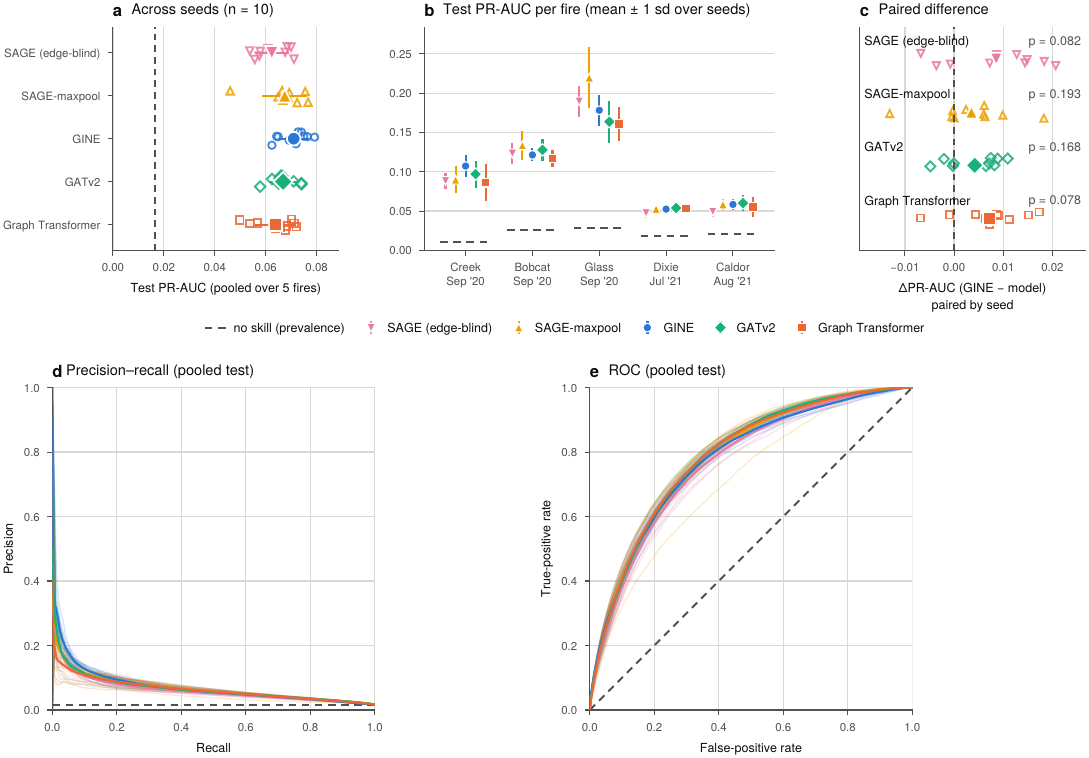} \caption{Sealed-test performance of the 5 \FGN{} variants over 10 seeds. (a) Pooled test PR-AUC per seed, mean $\pm$ 1 sd; dashed line, no skill (prevalence 0.0165). (b) Per-fire PR-AUC, mean $\pm$ 1 sd over seeds; dashes, per-fire prevalence. (c) Per-seed paired differences against the GINE variant with Holm-corrected Wilcoxon $p$-values. (d, e) Pooled precision--recall and ROC curves: thin lines, seeds; thick lines, seed mean; dashed, no skill.} \numlabel{figure}{fig:edsealed} \end{figure*}
\begin{figure*}[!tp] \centering \includegraphics[width=0.84\textwidth]{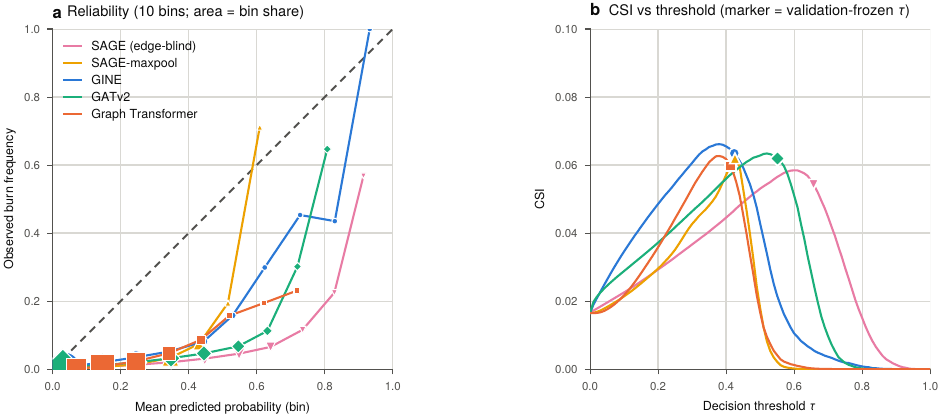} \caption{Operating behavior on the sealed test (10-seed pooled raw predictions, selected configurations). (a) Reliability diagrams; marker area $\propto$ bin population. (b) CSI against the decision threshold $\tau$; marker, the validation-frozen $\tau$. The variants differ in the width of this plateau rather than in its peak height. Brier score, ECE and plateau width for each model, the CSI cost of overshooting the threshold and the development-pool replication are in Supplementary Table~\ref{tab:calibration}.} \numlabel{figure}{fig:edcalibration} \end{figure*}
\begin{table*}[t]
\centering
\caption{Track-age-banded test AP (10-seed means) and the Spearman correlation of
snapshot-level AP with track age ($\rho$, computed on the 10-seed mean prediction; all $p < 10^{-13}$, $n = 322$
scoreable snapshots, i.e., those with at least one positive candidate). Bottom: the population weighting that
makes pooled AP an old-fire-dominated statistic.}
\label{tab:age}
\small
\setlength{\tabcolsep}{4pt}
\begin{tabular}{lcccccc}
\toprule
Model & 0--48\,h & 48--120\,h & 120--240\,h & 240--480\,h & 480\,h+ & $\rho$(age, AP) \\
\midrule
SAGE (edge-blind) & 0.434 & 0.195 & 0.154 & 0.044 & 0.033 & $-0.46$ \\
SAGE-maxpool & 0.436 & 0.228 & 0.168 & 0.056 & 0.040 & $-0.47$ \\
GINE & 0.481 & 0.206 & 0.181 & 0.049 & 0.038 & $-0.45$ \\
GATv2 & 0.411 & 0.222 & 0.166 & 0.046 & 0.036 & $-0.41$ \\
Graph Transformer & 0.388 & 0.198 & 0.166 & 0.050 & 0.036 & $-0.43$ \\
MLP (no message passing) & 0.330 & 0.157 & 0.074 & 0.021 & 0.022 & $-0.41$ \\
Null: ring-distance & 0.247 & 0.135 & 0.112 & 0.048 & 0.026 & $-0.54$ \\
\midrule
Share of candidates & 0.9\% & 2.6\% & 5.6\% & 13.1\% & 77.8\% & \\
Share of positives & 8.3\% & 11.3\% & 15.1\% & 13.7\% & 51.6\% & \\
\bottomrule
\end{tabular}
\end{table*}
\numlabelafter{table}{tab:edage}
\begin{table*}[t]
\centering
\caption{Weather ablation: per-snapshot AP delta of the
weather-masked arm minus the full model (positive = better \emph{without}
weather), paired over 11 leave-one-fire-out (LOFO) folds $\times$ 5 seeds. Depths 3--5 degrade significantly when the node weather and the wind edge channel are both masked, and depths 3--4 when only the node weather is masked (Holm $p < 0.05$; depth $p$-values are Holm-corrected within each arm, the pooled $p$ is a single uncorrected test). The pooled gain and these losses are the pooled-metric reversal that Section~\ref{sec:results-structure} discusses.}
\label{tab:b12}
\small
\begin{tabular}{lccc}
\toprule
Stratum & $\Delta$AP & $p$ & $n$ pairs \\
\midrule
\multicolumn{4}{l}{\emph{node weather masked}} \\
\quad pooled, all depths & $+0.0096$ & 0.0139 & 467 \\
\quad depth 1 & $+0.0057$ & 0.1439 & 467 \\
\quad depth 2 & $-0.0057$ & 0.1439 & 255 \\
\quad depth 3 & $-0.0215$ & 0.0008 & 131 \\
\quad depth 4 & $-0.0431$ & 0.0140 & 83 \\
\quad depth 5 & $-0.0259$ & 0.0616 & 61 \\
\quad per-fire & \multicolumn{3}{l}{9/11 fires degrade without weather} \\
\multicolumn{4}{l}{\emph{node weather and wind edge channel masked}} \\
\quad pooled, all depths & $+0.0152$ & $2.7\times 10^{-5}$ & 467 \\
\quad depth 1 & $+0.0122$ & 0.0024 & 467 \\
\quad depth 2 & $-0.0094$ & 0.1814 & 255 \\
\quad depth 3 & $-0.0276$ & 0.0307 & 131 \\
\quad depth 4 & $-0.0446$ & 0.0307 & 83 \\
\quad depth 5 & $-0.0426$ & 0.0031 & 61 \\
\quad per-fire & \multicolumn{3}{l}{10/11 fires degrade without weather} \\
\bottomrule
\end{tabular}
\end{table*}
\numlabelafter{table}{tab:edweather}
\begin{table*}[t]
\centering
\caption{Pre-registered expansion experiments, all paired against the same
locked-GINE control (11-fold leave-one-fire-out $\times$ 5 seeds). Micro =
pooled per-snapshot paired Wilcoxon (seed-averaged AP, $n = 467$ snapshot
pairs); fire-level = mean per-fire pooled-AP delta, count of fires improved, and a one-sample Wilcoxon $p$ over the 11 per-fire deltas.
Win rate is the share of paired snapshots on which the treatment arm scores higher. FiLM, feature-wise linear modulation; FRP, fire radiative power. Driver-type strata are in Supplementary Note~\ref{supp:enrich}.}
\label{tab:negatives}
\small
\setlength{\tabcolsep}{4pt}
\begin{tabular}{lcccccc}
\toprule
 & \multicolumn{3}{c}{Micro (per-snapshot)} & \multicolumn{3}{c}{Fire-level (macro)} \\
\cmidrule(lr){2-4} \cmidrule(lr){5-7}
Treatment & $\Delta$AP & $p$ & win rate & mean $\Delta$ & improved & $p$ \\
\midrule
Driver-label conditioning (feature) & $+0.0032$ & 0.0037 & 0.565 & $-0.0080$ & 5/11 & 0.58 \\
Driver-label conditioning (FiLM) & $+0.0004$ & 0.25 & 0.525 & $-0.0058$ & 4/11 & 0.17 \\
FRP intensity auxiliary head & $-0.0020$ & 0.71 & 0.507 & $-0.0178$ & 3/11 & 0.12 \\
24\,h auxiliary head (joint training) & $+0.0059$ & 0.00056 & 0.561 & $-0.0153$ & 4/11 & 0.46 \\
\bottomrule
\end{tabular}
\end{table*}
\numlabelafter{table}{tab:ednegatives}
\begin{figure*}[!tp] \centering \includegraphics[width=\textwidth]{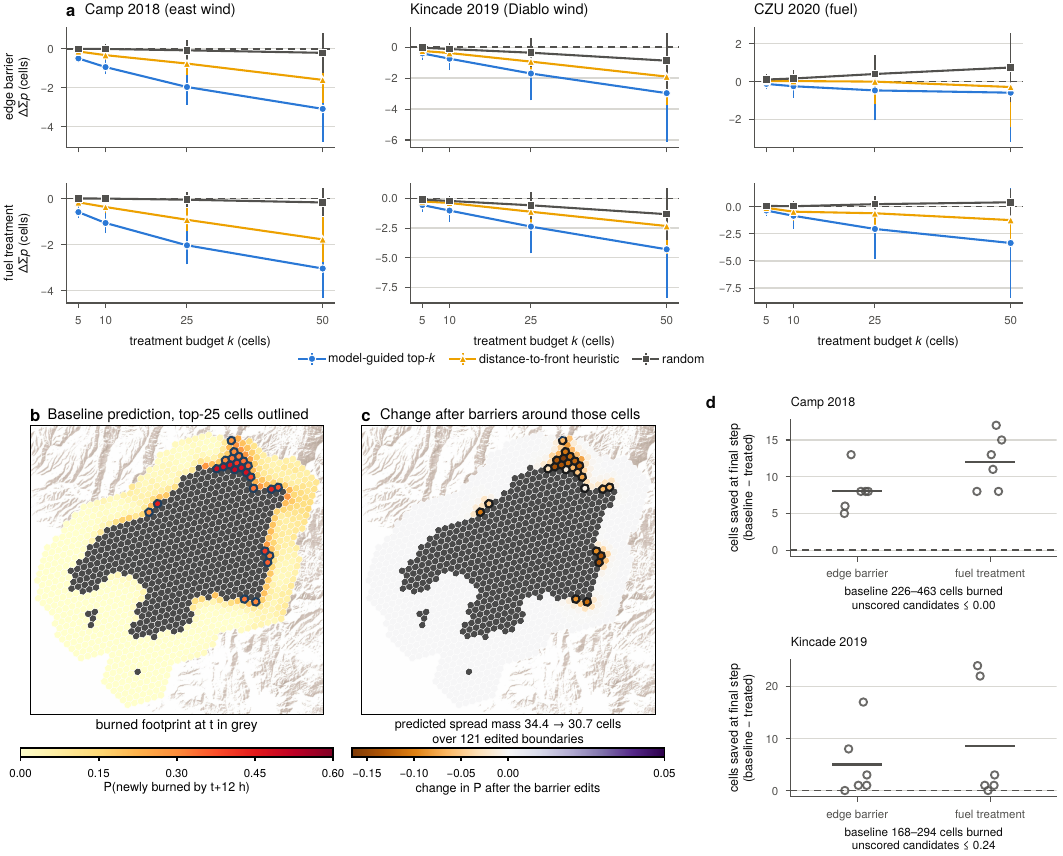} \caption{Intervention counterfactuals on held-out fires. (a) Dose--response (frozen fold models, 3 seeds; mean $\pm$ 1 sd over records): change in predicted spread mass $\Delta\Sigma p$ against treatment budget $k$, for edge barriers (top) and fuel treatments (bottom), under model-guided top-$k$, distance-to-front and random placement. On the fuel-dominated CZU, random placement \emph{increases} predicted spread for both intervention types (Supplementary Note~\ref{supp:interventions}). (b, c) One barrier counterfactual on Camp Fire, 9 November 2018, 12:00 UTC, from a frozen fold model (GINE, seed 0). Panel b shows the baseline probability of burning by $t + 12$\,h with the 25 highest-ranked cells outlined, and panel c the change after marking every edge of those cells as a barrier, which sets the barrier feature to 1 and fuel continuity to 0. Supplementary Note~\ref{supp:interventions} gives the numbers. (d) Autoregressive rollouts (3 steps of 12\,h, $k = 25$ placement at step 0): cells saved at the final step (baseline $-$ treated), one open marker per start time and seed, bar = mean. Headers give the baseline final burned-cell range and the maximum fraction of candidate cells that fell outside the fire's feature universe and could not be scored (0.24 on Kincade, 0.00 on Camp). All panels show sensitivities of a learned model rather than estimates of suppression effect. Basemap tiles: Esri World Terrain.} \numlabel{figure}{fig:edinterventions} \end{figure*}
\clearpage
\twocolumn %
\expandafter\let\csname figure*\endcsname\origfigstar \expandafter\let\csname endfigure*\endcsname\origendfigstar
\expandafter\let\csname table*\endcsname\origtabstar \expandafter\let\csname endtable*\endcsname\origendtabstar
\SuppTablenumbering
\section*{Supplementary Information}
\suppnotesection{Extended review of related work}\label{sec:related}

Learned models are trained on past fires rather than built from fire physics, and most published spread models of this kind take raster inputs, i.e., gridded maps of fire masks and environmental variables, which is also the form in which the dominant public datasets are released \citep{andrianarivony2024, huot2022, gerard2023}. Convolutional neural network (CNN) formulations, which are image models trained on such rasters, reflect the recent state-of-the-art architecture. Their reported F1-scores of 0.94--0.97 come, for example, from the FirePred hybrid CNN trained on British Columbia fires and evaluated on fires in Alaska and Nova Scotia, and from a CNN with atrous spatial pyramid pooling on the public Next Day Wildfire Spread (NDWS) dataset under a modified $7\times7$ neighborhood setup \citep{andrianarivony2024}. Public benchmarks such as NDWS and WildfireSpreadTS (WSTS) pose the task as next-day raster segmentation \citep{huot2022, gerard2023}, and how their scores depend on the evaluation setup is examined below. Newer formulations are also emerging, namely ranked point-set prediction \citep{bai2026wisp}, diffusion surrogates \citep{yu2026diffusion}, and graph formulations in which the landscape is a set of points or cells joined to their neighbors \citep{jiang2022, roesch2024}, so that spread becomes a process on the cells or on the joins between them.
\paragraph*{Learned wildfire spread prediction.}  \citet{andrianarivony2024} systematically review machine and deep learning for wildfire spread prediction. They organize the literature into CNNs, convolutional recurrent networks (CRNs) and time-series models, transformers, reinforcement learning, and graph neural networks, and conclude that CNNs and CRNs have achieved the best results to date. Besides the FirePred CNN and the atrous-pooling CNN on Next Day Wildfire Spread described above, the strongest results it compiles include a self-attentive ConvLSTM. Compiled accuracies reach 98.6\% (a multi-kernel CNN on the same benchmark). However, the same review lists F1-scores of 0.36 and 0.71 for that same benchmark, and cautions that comparison across papers is difficult because datasets and metrics differ. This suggests that the attainable score is set largely by the evaluation construction, which is a premise we test directly on our own task (Supplementary Note~\ref{sec:v1}). %

  Among the newer formulations, \citet{bai2026wisp} recast next-day active-fire forecasting as ranked point-set prediction, where a query-based model predicts active-fire cluster centers on a 375\,m grid across globally distributed regions (38.2\% average precision), and construct a global multi-source benchmark for the task. Denoising-diffusion surrogates likewise produce ensemble spread forecasts \citep{yu2026diffusion}. \citet{esparza2026modular} make the physics inside raster models inspectable, in that they add wind-conditioned and slope-conditioned attention biases, physics-feature retrieval and fire-conditioned gating to five backbones on Next Day Wildfire Spread. They find that the constructed wind bias aligns closely with the input wind but much more weakly with the observed next-day fire displacement. %
\paragraph*{Physics-based simulators.} Alongside learned prediction, physics-based simulators remain the operational reference. FARSITE, built by the US Forest Service on the Rothermel surface-spread model, and Canada's Prometheus are robust and validated through years of field use. However, they require precise inputs and extensive data preparation, and they struggle with real-time adaptability because of their computational demands \citep{andrianarivony2024}. National systems play the same role elsewhere, such as PHOENIX RapidFire and SPARK in Australia \citep{singh2025aureview}. These simulators also remain competitive with learned models, since on the 2023 Maui fires FARSITE delivered higher precision and F1 than the best of 5 deep-learning models \citep{kim2025maui}. Furthermore, when burn-probability maps derived from simulators and from machine learning disagree spatially, \citet{costasaura2023} conclude that the two families answer different questions.
\paragraph*{The sub-daily gap and label provenance.} The widely used public benchmarks built from observed satellite fire data, Next Day Wildfire Spread and WildfireSpreadTS, are daily, so models are trained and ranked on daily targets. At sub-daily granularity, the observation literature offers reconstruction rather than forecasting, in that FEDS assembles VIIRS active-fire detections into object-tracked perimeter sequences at roughly 12-hour cadence, which are growth histories \citep{chen2022feds,feds_zenodo}. Learned sub-daily models have mostly been benchmarked on simulations. FireBench, for example, poses spread prediction at one-second steps on large-eddy simulations as a public machine-learning competition \citep{ho2026firebench}. The DCIGN compiled in the review predicts 6\,h burn maps and is trained and tested on fires generated by the Rothermel model and by FARSITE \citep{andrianarivony2024}. The diffusion surrogate above trains on cellular-automata simulations with 2\,h snapshots subsampled to 20\,h frames \citep{yu2026diffusion}. The irregular-graph simulator of \citet{jiang2022} trains on 700 FARSITE-simulated one-hour fires, with a single real fire (the 2019 Getty Fire) as its observational test. Two studies do forecast at sub-daily steps against observed data, but their target is satellite fire activity rather than perimeter growth. FIRA predicts hourly gridded active-fire and fire radiative power fields against the observation-based GOES/VIIRS RAVE product \citep{hung2025fira}, and \citet{kim2025maui} forecast future VIIRS detection rasters at 12--48\,h steps. The FIRE-AID project also reports sub-daily spread prediction initialized from satellite-detected fire perimeters, using a hybrid architecture that includes graph-based message passing to forecast up to 12\,h ahead \citep{laux2026fireaid, esa2026fireaid}; its conference abstract and project report, however, do not set out an evaluation protocol comparable to the one used here. Neither reports quantitative results, and the model is trained on a partly proprietary satellite archive. The project also aims to couple the model with a reinforcement-learning recommender for firebreak placement \citep{laux2026fireaid}. Until very recently, no public dataset posed sub-daily spread forecasting against observed fire perimeters. PyroStack now provides public 12\,h FEDS-MTBS fire progression (MTBS, Monitoring Trends in Burn Severity) for 6{,}994 US fires from 2012 to 2024 together with hourly drivers. It defines a $t + 12$\,h fire-area target that includes the area already burned \citep{kondur2026pyrostack}. Its baseline, a vision transformer, is shown only qualitatively against observed perimeters on one held-out fire, and the work is a preprint at the time of writing. Our 12\,h task is possible because the FEDS archive makes the 12\,h response-planning window itself a supervised target. The same VIIRS lineage underlies the fire radiative power (FRP) fields used as the auxiliary target in Supplementary Note~\ref{supp:enrich}, where this shared provenance is disclosed. %
\paragraph*{Graph formulations.} Graph and cell representations of fire growth predate learning on graphs. Cellular growth simulators propagate fire over regular lattices, such as Cell2Fire at strategic-planning scale \citep{pais2021cell2fire}. \citet{jiang2022} move off the lattice with an irregular graph built by Delaunay triangulation and refined until each edge is homogeneous in fuel and slope. A per-edge neural regressor and a minimum-travel-time pass then simulate spread with explicit node-to-node routes. None of this lineage uses message passing, and \citet{jiang2022} themselves name graph convolutional networks as future work.

The closest prior work is \citet{roesch2024}, who formulate wildfire spread as node classification on an H3 hexagonal graph (resolution~9, about 350\,m across) over daily satellite-derived burned-area perimeters for 3{,}020 Mediterranean fires. They use a single spatiotemporal graph convolutional network with a gated recurrent unit (GCN--GRU) over undirected, unweighted, feature-less edges. We build on this formulation and extend it along the axes their study leaves open. The first is a \textbf{sub-daily (12\,h) horizon} from VIIRS-based perimeter tracking. The second is \textbf{directed graphs with typed directional edge features} (per-edge wind alignment, slope, fuel continuity, barriers), whereas their model encodes wind direction only as a binned scalar node feature. The third is the \textbf{systematic architecture comparison} (5 message-passing architectures plus a no-message-passing control) that they explicitly identify as future work. The fourth is an evaluation protocol designed for prediction-time validity, since their scored universe is derived from the fire's \emph{final} perimeter, and our own first protocol shared this hindsight construction in a different form, which we replaced after finding that it inflated the measured skill (Supplementary Note~\ref{sec:v1}). Their inputs include VIIRS hotspot counts and fire radiative power aggregated over day $t_n$, while the burned state is taken on the morning of $t_n$, so the afternoon detections of $t_n$ fall inside the interval the target describes. Furthermore, their stratified-random split is drawn over overlapping 5-day windows. Specifically, the 3{,}020 fire series become 11{,}082 windows at a 1-day step, and it is these windows that are split 7{,}686 / 1{,}736 / 1{,}660 (their Table 2). So windows sharing four of five days of one fire can fall in both train and test. We therefore do not compare our numbers to their reported weighted macro-mean IoU (0.36--0.37). That score averages the positive and negative classes at $\sim$94\% negative prevalence (implied by their class weights) and is not a positive-class skill measure, and no null baseline or PR analysis is reported against which to place it.

Beyond this closest neighbor, \citet{das2025compsac} compare 3 message-passing GNNs (GCN, the graph attention network GAT, and GraphSAGE) against CNN segmentation baselines for predicting the final burned area resulting from an ignition, on purely spatial graphs. Their best model, a GCN, outperforms the CNN baselines with an AUPRC of 0.479. A recent conformal-risk-control study of wildfire-spread prediction for evacuation mapping also evaluates tabular, spatial, and graph-based models side by side \citep{anon2026arxiv}. Graph neural networks have also been applied to wildfire impacts rather than spread. \citet{esparza2026graphfire} represent a community as a directed contagion graph whose edges carry physics-informed convection, radiation and ember terms, and predict building damage across six fires in California, Colorado and Canada. %

Outside fire, spatiotemporal graph convolutional networks have been used for short-horizon hazard nowcasting, both for urban flood inundation at the census-tract level \citep{farahmand2023flood} and for the flooding status of road segments 2--4 hours ahead \citep{yuan2022road}. %
\paragraph*{Evaluation protocols and benchmark design.}  Next Day Wildfire Spread \citep{huot2022} predicts tomorrow's fire mask on 1\,km fire-centered patches under a random-by-week split, where a persistence baseline captures a large share of the best reported AUC-PR (Section~\ref{sec:nulls}). WildfireSpreadTS \citep{gerard2023} and its multi-year extension WSTS+ \citep{lahrichi2025} predict next-day active fire on 375\,m tiles under leave-year-out evaluation, with a similarly strong persistence baseline. Mesogeos \citep{kondylatos2023} poses next-day danger forecasting and final burned-area prediction from ignition rather than a spread task. Three conventions of this family matter for what follows. The first is that the prediction target includes persisting fire, so a persistence baseline captures much of the attainable score. The second is that models are trained on the same persistence-containing targets. The third is that headline results are pooled over the test population without per-fire strata. We evaluate these conventions directly, with a persistence-containing objective tested in the raster head-to-head (Section~\ref{sec:results-raster}) and pooled-only reporting examined throughout Section~\ref{sec:results}. Recent work on the same benchmarks shows that the choice of metric can reorder the models \citep{gopakumar2026wsbench} and that standard thresholds on NDWS capture only 7--72\% of true fire spread \citep{anon2026arxiv}. The calibration of a next-day raster model has also been measured with the Brier score and ECE \citep{chakravarty2025uq}.  None of these compares graph and raster formulations trained on new growth only, or tests whether a decision threshold fixed on past fires transfers to held-out fires, which is the comparison of Section~\ref{sec:results-raster}. %

Such construction effects are beginning to be measured inside the raster family itself. On Next Day Wildfire Spread, \citet{firesensenet} show that two common protocol shortcuts, scoring the previous-day fire mask as part of the target and relabeling unknown pixels, inflate reported F1 by 44--50\% across the architectures they test.

Beyond fire science, our protocol draws on documented lessons from other fields. Strong baselines can reorder deep-learning claims when evaluation is careful \citep{grinsztajn2022}, and operationally framed geoscience benchmarks pair learned models with deterministic climatological baselines for the same reason \citep{mouatadid2023subseasonal}. Furthermore, recent evaluation guidance for machine-learned earth-system models calls for independent evaluation and comprehensive metric disclosure to guard against cherry-picking \citep{ullrich2025}. In the same spirit, we mechanically enforce the separation between model selection and the test set, and we report every contact with the test set.
\paragraph*{Fire drivers and wind events.}  \citet{abatzoglou2023} attribute a distinct, destructive population of western-US fires to downslope-wind events, which motivated the wind stratum we pre-registered for the out-of-state confirmation set. Daily cadence can also erase this signal, in that \citet{firesensenet} find that wind speed, conventionally a key fire-weather variable, acts as noise at the daily temporal resolution of the Next Day Wildfire Spread dataset. This is one more reason we pose the 12\,h horizon and the per-edge wind-alignment features together. %
\paragraph*{Fuel treatment and suppression modeling.} The intervention analysis of Section~\ref{sec:results-interventions} borrows its evaluation designs from the simulation literature, namely strategic-versus-random fuel-treatment placement from \citet{finney_gtr610}, paired treated-versus-untreated run deltas from the SimHarness evaluation of mitigation strategies \citep{simharness2023}, and learned firebreak placement on Cell2Fire from \citet{cell2fire_rl}. A learned surrogate that takes firebreaks as inputs has also been trained, although on simulated rather than observed spread \citep{zheng2026firebreaks}. There is now community-scale evidence for the placement effect on a graph as well. \citet{chulahwat2026} model urban conflagration with a validated graph-theory--based model in which vertices are ignitable fuels (buildings and vegetation) and edges carry the probability of ignition between them. For the Palisades area of Los Angeles, they find that combined vegetation management and home-ignition-zone hardening reduces mean relative building vulnerability by up to 57\% under random placement, and by up to 78\% when placement targets the nodes' modeled fire transmissibility. Suppression activity itself is deliberately absent from our model inputs, for the reason given in Supplementary Note~\ref{supp:interventions}, and the intervention probes are therefore disclosed as counterfactual sensitivities of a learned model rather than estimates of suppression effect (Section~\ref{sec:results-interventions}). A quality-assured archive of constructed firelines with engagement outcomes \citep{arkowitz2025} has recently become available. We do not use it here, but it is the record against which future placement experiments could be evaluated.

\suppnotesection{Our first evaluation setup and why it was replaced}\label{sec:v1} Our first-generation protocol built its candidates from the realized spread, which is akin to how the final-perimeter-conditioned evaluation of \citet{roesch2024} builds its scored cells: positives were flooded outward through the \emph{realized} new-burn region to any depth, while negatives were generated only adjacent to the realized advance. So which cells became candidates at all depended on where the fire actually went. Everything in this paper uses the constructible protocol of Section~\ref{sec:data}, and the first protocol is described here only to show why it had to be replaced. %

Under the first protocol's candidate construction, a hindsight hop-distance ranking that uses no features at all recovers $\sim$88\% of the best tuned GNN's average precision (0.306 vs 0.346 $\pm$ 0.021), because the candidate set itself encoded the outcome. Deep-ring candidates existed almost only where the fire actually ran ($\geq$94\% of deep rings unlabeled; labeled deep base rates 0.32--0.65 against true ring rates 0.006--0.018). Models trained on this distribution learn inverted physics and cannot be deployed on any prediction-time candidate set.

This failure mode, in which the candidate construction rather than the model carries the signal, motivates every protocol element of Section~\ref{sec:data}.
\begin{table*}[t]
\centering
\caption{Per-fire sealed-test PR-AUC (10-seed means, tie-aware AP).
Prev.\ = per-fire positive prevalence. CNN-M = objective-matched Res18-U-Net;
CNN-F = full-mask Res18-U-Net; Null = ring-distance (near-first). The per-fire structure is analyzed in Supplementary Note~\ref{app:mechanism}.}
\label{tab:perfire}
\small
\setlength{\tabcolsep}{4pt}
\begin{tabular}{lccccccccc}
\toprule
Fire & Prev. & SAGE & SAGE-maxpool & GINE & GATv2 & Transformer & CNN-M & CNN-F & Null \\
\midrule
Creek 2020 & 0.0109 & 0.0883 & 0.0897 & 0.1073 & 0.0970 & 0.0862 & 0.1210 & 0.0145 & 0.0262 \\
Bobcat 2020 & 0.0260 & 0.1231 & 0.1336 & 0.1214 & 0.1278 & 0.1167 & 0.0924 & 0.0459 & 0.0634 \\
Glass 2020 & 0.0283 & 0.1895 & 0.2197 & 0.1782 & 0.1636 & 0.1607 & 0.2075 & 0.0767 & 0.0684 \\
Dixie 2021 & 0.0180 & 0.0475 & 0.0527 & 0.0524 & 0.0539 & 0.0531 & 0.0682 & 0.0241 & 0.0379 \\
Caldor 2021 & 0.0211 & 0.0492 & 0.0583 & 0.0584 & 0.0601 & 0.0548 & 0.0958 & 0.0415 & 0.0511 \\
\bottomrule
\end{tabular}
\end{table*}
\numlabelafter{table}{tab:edperfire}
\begin{table*}[t]
\centering
\caption{Operating behavior of the six models on the sealed test and in its development-pool replication. The left block gives the sealed test (10 seeds, selected configurations, raw
probabilities), namely the Brier score, ECE, the width of the $\tau$ plateau holding
$\geq 90\%$ of CSI$_{\max}$, and the CSI cost of overshooting the selected
threshold $\tau$ by $+0.05$. The right block gives the out-of-sample replication on the 11-fire
development pool at the \emph{selected} configurations (macro over held-out
fires, 3 seeds).}
\label{tab:calibration}
\small
\setlength{\tabcolsep}{4pt}
\begin{tabular}{lcccccc}
\toprule
 & \multicolumn{4}{c}{Sealed test (selected)} & \multicolumn{2}{c}{Dev pool (selected)} \\
\cmidrule(lr){2-5} \cmidrule(lr){6-7}
Model & Brier & ECE & Plateau & CSI @ $\tau{+}0.05$ & Brier & ECE \\
\midrule
SAGE (edge-blind) & 0.094 & 0.220 & 0.157 & $-7\%$ & 0.071 & 0.158 \\
SAGE-maxpool & 0.064 & 0.191 & 0.052 & $-49\%$ & 0.053 & 0.142 \\
GINE & 0.035 & 0.093 & 0.143 & $-7\%$ & 0.029 & 0.047 \\
GATv2 & 0.060 & 0.147 & 0.157 & $-9\%$ & 0.040 & 0.085 \\
Graph Transformer & 0.050 & 0.163 & 0.099 & $-16\%$ & 0.041 & 0.115 \\
MLP (no message passing) & 0.093 & 0.271 & 0.074 & $-62\%$ & 0.069 & 0.196 \\
\bottomrule
\end{tabular}
\end{table*}

\suppnotesection{Architecture mechanism: why the GNNs tie, and where architecture matters}\label{app:mechanism}

\paragraph*{Why the architectures tie on ranking metrics.}
 On the sealed test, under one tuning contract and identical inputs, 5 ways of combining neighborhood information land inside a single seed-noise band (Extended Data Figure~\ref{fig:edsealed}a--c, with overlapping precision--recall and ROC curves in its panels d and e). Three properties of the task setting explain the tie. The first is that most of the skill sits one hop away, since pooled skill is dominated by the ranking of cells adjacent to the front, and one hop of neighborhood context is a setting where any message-passing scheme can move the same information. The second is that the lattice has near-constant degree. On H3 hexagons (at most 6 neighbors), sum, mean, and attention aggregation differ far less than on irregular graphs, because there is no degree heterogeneity for normalization or attention to exploit. The third is that the features limit skill more than the routing does. The weather ablation moves depth-3--5 AP by $-0.028$ to $-0.045$ per stratum, which is several times more movement than any architecture swap, with a graded response through the wind \emph{edge channel}.  Together with the control results (the control at the null pooled, and below it beyond 240\,h track age), the architecture result is that message passing is necessary, the edge information matters, and the choice of aggregator does not measurably change ranking performance. Hence, the practical recommendation is to choose the architecture on operating behavior, i.e., calibration and threshold robustness. %

\paragraph*{Operating behavior, where the architectures differ.} On calibration and threshold robustness the architectures do separate (Supplementary Table~\ref{tab:calibration}; Extended Data Figure~\ref{fig:edcalibration}). Capacity does not explain the ordering, since the models span 58\,k to 682\,k parameters and the largest sits in the middle of the range. The mechanism does explain it, in that the miscalibration ordering is the overconfidence ordering. Sharpness (mean $|p - \text{prevalence}|$) rank-correlates with ECE at Spearman $\rho = 1.0$ and with Brier at 0.89, which is descriptive rather than inferential with $n = 6$ models. It spans 2.3$\times$ across architectures at tied AP (GINE 0.097 $<$ GATv2 0.152 $<$ Graph Transformer 0.163 $<$ SAGE-maxpool 0.191 $<$ SAGE 0.222, and 0.271 for the control). Discrimination does not require overconfidence, since GINE attains the third-best class separation (mean $p^{+} - p^{-} = 0.142$, behind SAGE 0.200 and GATv2 0.194) at the lowest sharpness. Two pathways set the confidence scale, although the causal links are our interpretation and the loss analysis reads the selection studies rather than the test. The first is the tuned focal loss, whose selected inflation depends on the aggregator. Within the 18-trial selection studies, GINE's selection score degrades as the loss becomes more confidence-inflating ($\rho(\text{value}, \gamma) = -0.72$, $p = 0.001$), while the edge-blind SAGE ranks well only under heavy positive upweighting by the focal-loss class weight ($\rho(\text{value}, \alpha) = +0.55$, $p = 0.02$). The second is the architecture itself at fixed loss, since the untuned control trains at SAGE-maxpool's selected loss (focal-loss $\alpha = 0.75$, $\gamma = 2$) and is still sharper (0.271 vs 0.191) with far worse separation.

An out-of-sample replication on the 11-fire development pool (leave-one-fire-out, selected configurations, 3 seeds) reproduces the \textbf{identical ranking} of all 6 models on ECE (Spearman $\rho = 1.0$). It also reproduces the Brier ranking up to one adjacent swap of GAT and the transformer ($\rho = 0.94$), with GINE again best-calibrated (Brier 0.029, ECE 0.047 macro-averaged over held-out fires). Hence, the calibration ordering is not an artifact of the single sealed-test snapshot.

\paragraph*{Architectures across track-age bands.} %
In the track-age bands of Extended Data Table~\ref{tab:edage}, the lift of the GINE variant over the null thins to parity in the 240--480\,h band. GAT and the edge-blind SAGE fall marginally below the null in that same band, while the no-message-passing control falls well \emph{below} the null beyond 240\,h, which implies that message passing carries the skill on old fires.

\paragraph*{Exploratory structure: the fire--architecture interaction.}The per-fire structure behind the headline tie was analyzed post hoc from saved predictions only. The development and sealed-test pools were never mixed inside a statistic and cross-pool sign agreement was used as the replication filter. A two-way ANOVA on per-fire AP puts fire identity at 78.8\% (development pool) and 87.3\% (test pool) of variance, model choice at 1.5\% and 1.1\%, and the fire$\times$model interaction at 7.8\% ($p = 0.009$) and 3.2\% ($p = 2.5\times10^{-9}$). Seed-split winner replication (chance $\approx 0.20$) averages 0.55 on the development pool and 0.52 on test, so roughly half of the per-fire ``winners'' are seed noise. All GNN pairs agree on which snapshots are hard ($\rho = 0.85$--$0.91$ on the development pool and 0.92--0.97 on test), so the architectures tie because small complementary strengths average out over a heterogeneous fire population rather than because they are interchangeable. Single-pool $p$-values here are not multiplicity-safe (19 fire characteristics $\times$ 5 models at $n = 11$), so we rely on cross-pool sign replication.

\suppnotesection{Enrichment experiments in full}\label{supp:enrich}
We pre-registered 3 expansion hypotheses on the development pool, each paired against the same locked-GINE control on identical folds, seeds, and compute backend, and each with a committed primary test and a fire-level cross-check. All three are negative (Extended Data Table~\ref{tab:ednegatives}). We report them in full, since each pooled test is well-powered, and two of the three would have been publishable ``wins'' under pooled-only reporting, which is the pattern the weather ablation of Extended Data Table~\ref{tab:edweather} also shows and Section~\ref{sec:discussion} takes up.

\paragraph*{Driver-type label.}\label{sec:driverlabel} %
The conditioning experiment of Section~\ref{sec:results-negatives} uses a snapshot-level binary label, \emph{wind-dominated} vs \emph{fuel-dominated} in the taxonomy of \citet{keeley2019}, computed from the weather at $t$, so we describe it here. It is the union of a synoptic downslope-wind detector, which implements the published thresholds of \citet{abatzoglou2021} on ERA5 pressure-level fields \citep{hersbach2020era5} with documented simplifications of the terrain geometry, and an HRRR gate on the maximum wind over the front cells ($\geq$8\,m\,s$^{-1}$ at $t$). The detector uses only the trailing 24\,h of ERA5 analyses (at least 2 qualifying 6-hourly analyses) and the gate uses the wind at $t$, so the label is computable at prediction time. Six Saddleridge pre-ignition snapshots predate its ERA5 window and take the HRRR gate alone. We validated the detector against published event records. It reproduces the downslope-windstorm case that \citet{abatzoglou2021} present (the 8 November 2018 Camp Fire) and flags the documented 27 October 2019 Kincade Diablo windstorm \citep{metz2023}. Woolsey is a documented miss at the synoptic scale, since its coastal Santa Ana gap flow is below ERA5's effective resolution, but the HRRR gate covers it (thresholds were kept as published). Wind-dominated snapshots make up 14.0\% of the corpus.

\paragraph*{Input-derived driver-type conditioning: negative, and informative.} The driver-type label is built from the weather at $t$ as described above. A pre-registered gate then established that the label marks a real weakness. Raw per-snapshot AP is indistinguishable between the two classes because wind snapshots carry 2.4$\times$ the base rate, but the prevalence-controlled lift (AP over prevalence) shows that the model captures proportionally less of the achievable ranking on wind snapshots (median lift 6.3 vs 13.0, Mann--Whitney $p = 4.6\times10^{-5}$). The fire-level version of this gate is underpowered, which is a caveat we carried into the pre-registration. Conditioning the locked trunk on the label, either as an appended feature or via feature-wise linear modulation (FiLM) \citep{perez2018film}, does not close the deficit. The feature variant's pooled micro win ($+0.0032$, $p = 3.7\times10^{-3}$) fails the fire-level cross-check (5/11 fires, mean $-0.0080$, $p = 0.58$), and the pre-registered per-class strata show where the pooled number comes from. The gain is entirely in the fuel class ($+0.0058$, $p = 3.1\times 10^{-5}$), while the wind class, which is the one the label was built for, gets significantly \emph{worse} ($-0.0143$, win rate 0.34, $p = 0.011$; FiLM $-0.0098$, $p = 0.013$). The gate's lift deficit also \emph{widened} under treatment (median fuel$-$wind gap 6.76 $\to$ 8.20 with the feature variant and 7.28 with FiLM). Hence, a time-$t$-computable driver-type label is a valid input that marks a real skill deficit, but snapshot-level binary conditioning at this capacity does not close it, and instead trades wind-class accuracy for fuel-class accuracy. This is consistent with the weather ablation, in that the wind information that helps is already flowing through the wind edge channel, and the binary label adds no usable structure beyond it.

\paragraph*{Fire-intensity auxiliary supervision: negative.} Fire intensity and spread are coupled aspects of the same process and are observed by the same sensor. Yet, to our knowledge, whether auxiliary intensity supervision improves a perimeter-spread model has not been tested. Hence, we added a second output head on the shared trunk that predicts the snapshot-level mean fire radiative power (FRP) over the same 12-hour window the spread label covers, trained with a masked Huber loss on log-FRP and with the loss weight selected on inner validation only. Adding this head did not improve spread prediction. The pooled per-snapshot change in AP was $-0.0020$ ($p = 0.71$, win rate 0.507), the change was negative in every depth stratum without reaching significance in any, and only 3 of 11 fires improved. Two caveats apply to this result. The FRP target shares its VIIRS lineage with the spread labels themselves \citep{chen2022feds}, and it is defined per snapshot rather than per cell, so per-cell FRP supervision remains untested future work. %

\paragraph*{A 24\,h auxiliary head: mixed.} Here we trained a second output head on the shared trunk to predict the 24-hour spread label alongside the 12-hour one, so that the smoother longer-horizon target could act as an extra training signal. The pre-registered primary test, which is the pooled per-snapshot comparison, favored the joint arm ($+0.0059$, $p = 5.6\times10^{-4}$, win rate 0.561). However, the fire-level cross-check reverses it, since 7 of 11 fires favor the 12-hour-only arm (mean $-0.0153$, $p = 0.46$). The two tests disagree because they weight the fires differently. The per-snapshot pool is dominated by the two fires with the most snapshots, August Complex and Carr--Hirz--Delta, whereas the fire with the largest loss, Saddleridge, contributes only 2 scoreable snapshots and so barely counts in the pooled test. The gain was also small in absolute terms, at $+0.0059$ on a base AP of 0.216, or about 2.7\% relative. We had expected the smoother 24-hour target to hurt the ranking of cells adjacent to the front, but this did not happen, and depth 1 was in fact the stratum with the largest positive change ($+0.0077$, Holm $p = 2.9\times10^{-5}$). Furthermore, the two heads almost never disagree on ordering, with the 24-hour probability falling below the 12-hour probability for only 0.2\% of candidates ($P(\hat p_{24} < \hat p_{12}) = 0.002$), so a variant that enforces this ordering through a survival-style head was not worth running. Lastly, the auxiliary head does learn its own target, reaching a median held-out 24-hour AP of 0.185 against a base rate of about 0.04, but this does not carry over into better 12-hour ranking.

\suppnotesection{Out-of-state pre-registration mechanics}\label{app:w2} The out-of-state set was declared evaluation-only at freeze time (SHA256 checksums written and verified before any evaluation). The single pass (the locked GINE variant at its selected configuration, the untuned no-message-passing control, and the ring-distance null, 10 seeds) was executed once under mechanical guards (sign-off environment variable, declared-pass flag, checksum verification, configuration guard), with interpretation commitments written down before the run. The transfer claim requires pooled lift $\gtrsim 1.5\times$ \emph{and} an established-track stratum whose lift also exceeds 1. The pre-registered per-fire, established-track, initial-window, driver-type and track-age strata are reported in Table~\ref{tab:w2}, and none of them replaces the pre-registered headline. The per-fire beyond-ring fractions are given in Section~\ref{sec:data}, and the per-ring-depth strata remain in the released records. The pre-registration document, which contains the design, the interpretation commitments, and the review checklist resolved before sign-off, is included verbatim in the released artifacts. %

\paragraph*{Out-of-state strata in detail.}The wind-dominated stratum holds its lift, at 1.52$\times$ against 1.35$\times$ for the fuel-dominated stratum (Section~\ref{sec:results-w2}). So the wind-class skill deficit of the driver-type gate (Supplementary Note~\ref{supp:enrich}) does not carry to the out-of-state set at the stratum level. This matters because the set is wind-heavy by design, and once the development-pool gate had found the deficit, the pre-registration named the wind stratum as the hard case. The initial-window stratum sits below its null (0.86$\times$) at high prevalence. While the in-state young-track advantage does not carry over (0--48\,h lift 1.01$\times$), the lift is largest on established tracks (1.73$\times$) and in the 480\,h+ band (3.78$\times$). The pooled 1.62$\times$ lift (Table~\ref{tab:w2}) is above 1 on every fire, which spans three fuel systems (Columbia Basin grass, beetle-killed Colorado lodgepole, Oregon Cascades timber). It is thin on East Troublesome, and it holds on Bootleg 2021, whose season postdates every training fire.

\begin{table*}[t]
\centering
\caption{Intervention counterfactuals on fires held out of their leave-one-fire-out fold models (Camp, Kincade and CZU; 3 seeds): change in predicted spread mass $\Delta\Sigma p$ (more negative =
larger predicted suppression) for model-guided top-$k$ placement across
budgets, versus the distance-to-front heuristic and random placement at $k = 25$.
Values are counterfactual sensitivities of a learned model rather than estimates of suppression effect. Random placement on CZU
\emph{increases} predicted spread for both intervention types (Supplementary Note~\ref{supp:interventions}).}
\label{tab:interventions}
\small
\setlength{\tabcolsep}{4pt}
\begin{tabular}{llcccccc}
\toprule
 & & \multicolumn{4}{c}{Top-$k$ (guided)} & Distance & Random \\
\cmidrule(lr){3-6} \cmidrule(lr){7-7} \cmidrule(lr){8-8}
Fire & Intervention & $k{=}5$ & $k{=}10$ & $k{=}25$ & $k{=}50$ & $k{=}25$ & $k{=}25$ \\
\midrule
Camp 2018 (east wind) & edge barrier & $-0.50$ & $-0.94$ & $-1.97$ & $-3.09$ & $-0.76$ & $-0.08$ \\
Camp 2018 (east wind) & fuel treatment & $-0.59$ & $-1.06$ & $-2.03$ & $-3.04$ & $-0.92$ & $-0.05$ \\
Kincade 2019 (Diablo wind) & edge barrier & $-0.40$ & $-0.75$ & $-1.70$ & $-2.97$ & $-0.93$ & $-0.36$ \\
Kincade 2019 (Diablo wind) & fuel treatment & $-0.56$ & $-1.02$ & $-2.37$ & $-4.29$ & $-1.12$ & $-0.58$ \\
CZU 2020 (fuel) & edge barrier & $-0.12$ & $-0.24$ & $-0.46$ & $-0.58$ & $+0.00$ & $+0.40$ \\
CZU 2020 (fuel) & fuel treatment & $-0.36$ & $-0.85$ & $-2.06$ & $-3.35$ & $-0.61$ & $+0.22$ \\
\bottomrule
\end{tabular}
\end{table*}

\suppnotesection{Sealed-test contact log}\label{app:contacts} The sealed in-state test set was scored five times: a 3-seed preliminary pass, the 10-seed finals, the full-mask raster-CNN pass, the objective-matched CNN pass, which was declared after the full-mask result had been read, and a pre-registered one-pass re-run of the 10-seed finals at the selected configurations after the configuration defect described in Supplementary Note~\ref{app:tuning} was found. The re-run supplies every number reported in this paper. No selection, hyperparameter choice or code change was conditioned on a test read, and all selection used development fires only. Numbers from the first protocol appear only in Supplementary Note~\ref{sec:v1}. The out-of-state confirmation set was scored exactly once, under a pre-registered protocol with mechanically enforced single-pass execution.
\suppnotesection{Intervention counterfactuals in full}\label{supp:interventions}
The edges and cells of \FGN{} carry the quantities that mitigation changes, which is what lets the graph formulation support a use beyond prediction. A fuel treatment swaps a cell's fuel inputs for those of a non-burnable cell from the same fire, and a barrier such as a road, a fuel break or a fireline changes the barrier and fuel-continuity features on the edges into and out of a cell. So we can ask the trained model what it would have predicted had those inputs been different, which is a counterfactual, i.e., a re-scoring of the unchanged model on edited inputs, rather than a simulation of fire physics. Specifically, we start from the model's own risk ranking on a held-out fire and place a treatment budget of $k$ cells either at the top-ranked cells, at random, or by distance to the front. We then edit only those input features and re-score the frozen model. Three patterns emerge on the fires held out of their fold models (Supplementary Table~\ref{tab:interventions}). First, for guided placements the reduction in predicted spread grows with the treatment budget, in every fire and for both intervention types (Extended Data Figure~\ref{fig:edinterventions}a). Second, model-guided placement dominates the alternatives. At a 25-cell budget it reduces the predicted spread mass 4--41$\times$ more than random placement on the wind-driven fires, and it beats a distance-to-front heuristic in all 6 fire$\times$intervention-type combinations, which reproduces the strategic-versus-random effect reported for physical simulators \citep{finney_gtr610}. Extended Data Figure~\ref{fig:edinterventions}b,c shows one Camp record at this budget, on 9 November 2018 at 12:00 UTC (GINE, seed 0). Marking every edge of the 25 top-ranked cells as a barrier, which sets the barrier feature to 1 and fuel continuity to 0 on 121 cell boundaries, lowers the predicted spread mass from 34.4 to 30.7 cells. The effect stays on the treated flank, and 18 cells rise slightly. It is the largest reduction among the 81 Camp records at this budget, whose mean is the $-1.97$ of Supplementary Table~\ref{tab:interventions}. Third, the directional signature is mixed. On Camp, wind-aligned barrier sets suppress predicted spread more than size-matched crosswind or upwind sets, as physics predicts, but the ordering inverts on Kincade and all directional placements are weak on CZU. The step-0 interventions also persist through 3-step (36\,h) rollouts (Extended Data Figure~\ref{fig:edinterventions}d). Before a fire season, the same model supports mitigation planning, because the quantities that mitigation changes are its inputs. So a planner can edit those inputs on past fire perimeters or ignition scenarios and re-score the frozen model to ask where a treatment would most reduce predicted spread. In other words, the model can be used to shortlist locations for fuel treatments and fuel breaks, so that the slower physics-based simulators only need to be run on the placements that make the shortlist.  %

We do note two caveats here. First, these are counterfactual sensitivities of a learned model rather than estimates of suppression effect, because the model is deliberately trained with no suppression information (crews are sent where there is most to protect, so suppression presence would encode the values at risk rather than fire behavior). Second, the probe also exposes where the model is not physical. On the fuel-dominated CZU, randomly placed barrier edits \emph{increase} predicted spread, which is a response to edited input configurations never seen in training rather than a learned association between barriers and fire. This limits simulator-style use of the raw model, and intervention-aware training is future work.

\suppnotesection{Choice of the 5-ring candidate horizon}\label{app:coverage}
The candidate horizon $R$ sets how far ahead of the current front a cell can be scored. A small $R$ misses deep runs, while a large $R$ floods the candidate set with cells that almost never burn. We chose $R = 5$ from a coverage analysis on the training and validation fires of the earlier corpus (coverage 79.7\% at $R = 3$, 88.5\% at $R = 5$, 96.3\% at $R = 10$), before any model under this protocol was tuned. Supplementary Table~\ref{tab:ringR} shows the same curve recomputed on the final corpus for the 11-fire development pool. Coverage rises quickly to $R = 3$ and then flattens, with each further ring adding 1.5--5.3 points of coverage up to $R = 8$ and about 1 point beyond, while adding about 130{,}000 candidates and pushing the base rate down.

Four considerations fixed $R = 5$. The first is that it is the largest radius the stored graphs support, since the node universe is built with a 5-hop buffer around the front, so any wider horizon needs a corpus rebuild and a feature fetch for the added cells. The second is that it matches the models' receptive field. The tuned architectures have 3 or 4 message-passing layers, so cells beyond the fourth ring already receive no message from a burned cell, and widening the horizon would add only such cells, which would receive no information from the front. The third is that the marginal return is poor, since extending $R = 5 \to 10$ raises development-pool coverage from 85.4\% to 93.4\% but doubles the candidate set (672\,k $\to$ 1.34\,M) and nearly halves the base rate (0.021 $\to$ 0.011). The fourth is that a dynamic, wind-scaled radius was tested and rejected, since the Spearman correlation between the frontier-mean HRRR wind at $t$ and the deepest advance in the window is 0.11 on the training and validation snapshots. So the wind at $t$ gives no usable signal for a per-snapshot radius. The cost of the choice is stated in Section~\ref{sec:data}, in that coverage is regime-skewed and much of the missing mass on the worst fires is window-0 detection latency that no radius recovers. A wider horizon paired with deeper or position-aware architectures is left as future work.
\begin{table}[H]
\centering
\caption{Coverage of realized 12\,h growth (fraction of newly burned cells within $R$ rings of the front) and candidate base rate as a function of the ring horizon $R$. The development pool is the 11 non-test fires, and the test column is descriptive only.}
\label{tab:ringR}
\small
\begin{tabular}{cccc}
\toprule
$R$ & Dev coverage & Dev base rate & Test coverage \\
\midrule
1 & 0.525 & 0.061 & 0.523 \\
2 & 0.683 & 0.040 & 0.700 \\
3 & 0.766 & 0.030 & 0.795 \\
4 & 0.819 & 0.024 & 0.851 \\
\textbf{5} & \textbf{0.854} & \textbf{0.021} & \textbf{0.888} \\
6 & 0.879 & 0.018 & 0.914 \\
7 & 0.897 & 0.015 & 0.931 \\
8 & 0.912 & 0.014 & 0.946 \\
9 & 0.924 & 0.012 & 0.959 \\
10 & 0.934 & 0.011 & 0.969 \\
\bottomrule
\end{tabular}
\end{table}
\begin{table*}[t]
\centering
\caption{The 16-fire / 909-graph corpus. Graphs are activity-gated $\sim$12\,h snapshots; the 120 retained zero-growth windows (48 in the test pool) and the pre-ignition snapshots of Bobcat 2020 and Saddleridge 2019 are described in Section~\ref{sec:corpus}. Candidates and positives are the
$R = 5$ constructible sets summed over each fire's snapshots, and cov@5 is the
fraction of realized 12\,h growth within five rings.
Hennessey (LNU) 2020 is outside the in-state split but is the 11th fire of
the development pool. Totals: 909 graphs, 1,184,772 candidates,
22,247 positives.}
\label{tab:corpus}
\small
\setlength{\tabcolsep}{4pt}
\begin{tabular}{llcccc}
\toprule
Fire & Role & Graphs & Candidates & Positives & cov@5 \\
\midrule
Camp 2018 & train & 27 & 22,062 & 539 & 0.92 \\
Carr--Hirz--Delta 2018 & train & 112 & 132,600 & 1,662 & 0.92 \\
Mendocino Complex 2018 & train & 59 & 81,826 & 2,244 & 0.94 \\
Woolsey 2018 & train & 11 & 7,049 & 244 & 0.63 \\
Kincade 2019 & train & 14 & 6,327 & 286 & 0.80 \\
Saddleridge 2019 & train & 10 & 1,580 & 12 & 1.00 \\
August Complex 2020 & train & 111 & 249,154 & 5,366 & 0.89 \\
Bobcat 2020 & test & 48 & 25,293 & 658 & 0.98 \\
Creek 2020 & test & 124 & 171,883 & 1,867 & 0.85 \\
CZU Complex 2020 & train & 16 & 8,525 & 335 & 0.72 \\
Dolan 2020 & val & 59 & 29,034 & 552 & 0.87 \\
Glass 2020 & test & 17 & 8,789 & 249 & 0.58 \\
Hennessey (LNU) 2020 & dev pool only & 24 & 41,850 & 1,303 & 0.80 \\
North Complex 2020 & val & 96 & 92,023 & 1,249 & 0.68 \\
Caldor 2021 & test & 63 & 54,652 & 1,151 & 0.90 \\
Dixie 2021 & test & 118 & 252,125 & 4,530 & 0.92 \\
\bottomrule
\end{tabular}
\end{table*}

\suppnotesection{Selection, tuning, and final configurations}\label{app:tuning} \label{sec:protocol} Hyperparameter selection for every variant ran as an 18-trial Optuna study \citep{akiba2019} (TPE sampler with successive-halving pruning) scored by mean leave-one-fire-out (LOFO) PR-AUC over the 10 in-state training and validation fires. The selected configuration was then retrained from scratch on 10 fresh seeds, early-stopped and thresholded ($\tau$) on the 2 validation fires, and evaluated on the pooled 5-fire sealed test. The raster baseline's selection used a 6-configuration sweep (learning rate $\times$ loss) on the same validation pair only, which is a comparable selection budget to the GNNs' 6 tuned dimensions and is disclosed as such. All selection, for every tuned model, used development fires only and never saw the test fires.

LOFO selection scores (0.173--0.203) sit roughly 3$\times$ above the pooled sealed-test scores (0.062--0.071). The two are different estimands, since selection averages over 10 development fires while the test pools 5 held-out fires dominated by 2 long-tailed megafires, and per-fire attainable AP varies three to four times across fires (Supplementary Table~\ref{tab:edperfire}), so absolute levels do not transfer across pools. The same caution applies in Section~\ref{sec:results-raster}. %

On the sealed test, the model with the best leave-one-fire-out score (GraphSAGE) was the worst of the variants (Section~\ref{sec:results-headline}). We report this as a caution against reading too much into the selection metric rather than as evidence of anti-correlation, since $n = 5$.

The first 10-seed finals (the second sealed-test contact) were trained, through a configuration defect since fixed, with the search's fixed architecture constants replaced by the repository's constructor defaults. The finals reported throughout this paper are the declared re-run at the selected configurations (the re-run, Supplementary Note~\ref{app:contacts}), and the first finals are kept only as a provenance record.
\suppnotesection{Pathway to operational testing}\label{supp:pathway}
Implementation should begin with a retrospective replay that reconstructs when each perimeter and environmental product actually became available, defines an explicit issuance time, and measures the remaining lead time to the next observed growth interval. Fire analysts and data providers should then establish quality checks for perimeter age, missing features, candidate coverage, and changing weather, accompanied by a wider-area assessment for growth beyond the five-ring domain. Before operational trials, the model should be calibrated on separate development data and tested on untouched seasons and regions using distance baselines, fire-level summaries, and decision thresholds selected with incident-management practitioners. A shadow deployment could present ranked candidate cells, data age, and the model's spatial scope alongside existing tools while measuring latency, false alarms, missed growth, and the usefulness of the output to analysts. Fuel-treatment and barrier edits should proceed through a separate validation track using independent simulators and, where suitable, documented fireline or treatment outcomes \citep{arkowitz2025}. Failures in input availability, transfer, calibration, or physical plausibility should trigger documented review and revision before predictions influence operational choices. This staged pathway makes the next step a testable integration with fire-management practice rather than an inference of readiness from retrospective ranking performance.

\end{document}